\documentclass[acmsmall]{acmart}

\AtBeginDocument{%
  \providecommand\BibTeX{{%
    \normalfont B\kern-0.5em{\scshape i\kern-0.25em b}\kern-0.8em\TeX}}}

\setcopyright{acmcopyright}
\copyrightyear{2020}
\acmYear{2020}
\acmDOI{10.1145/1122445.1122456}

\usepackage{multirow}
\usepackage{float}
\usepackage[section]{placeins}

\newcommand{\surveyUpdate}[1]{{#1}}

\begin{document}



\title[DL and XAI for Automatic Report Generation from Medical Images]{A Survey on Deep Learning and Explainability for Automatic Report Generation from Medical Images}

\author{Pablo Messina}
\authornote{Both authors contributed equally to this research.}
\email{pamessina@uc.cl}
\author{Pablo Pino}
\authornotemark[1]
\email{pdpino@uc.cl}
\author{Denis Parra}
\email{dparra@ing.puc.cl}
\author{Alvaro Soto}
\email{asoto@uc.cl}
\affiliation{%
  \institution{Computer Science Department, Pontificia Universidad Católica de Chile}
  \streetaddress{Vicuña Mackenna 4860}
  \postcode{7820436}
  \city{Santiago}
  \country{Chile}
}
\author{Cecilia Besa}
\email{besacecilia@gmail.com}
\author{Sergio Uribe}
\email{suribe@uc.cl}
\author{Marcelo Andía}
\email{meandia@uc.cl}
\affiliation{%
  \institution{Department of Radiology, School of Medicine, Pontificia Universidad Católica de Chile}
  \streetaddress{Avda. Libertador Bernando O'Higgins 340}
  \postcode{8320000}
  \city{Santiago}
  \country{Chile}
}
\author{Cristian Tejos}
\email{ctejos@puc.cl}
\affiliation{%
  \institution{Department of Electrical Engineering, Pontificia Universidad Católica de Chile}
  \streetaddress{Vicuña Mackenna 4860}
  \postcode{7820436}
  \city{Santiago}
  \country{Chile}
}
\author{Claudia Prieto}
\email{cdprieto@gmail.com}
\affiliation{%
  \institution{School of Biomedical Engineering and Imaging Sciences, King’s College London, St Thomas’ Hospital}
  \streetaddress{Lambeth Palace Rd}
  \postcode{SE1 7EH}
  \city{London}
  \country{UK}
}
\author{Daniel Capurro}
\email{dcapurro@unimelb.edu.au}
\affiliation{%
  \institution{School of Computing and Information Systems, The University of Melbourne}
  \streetaddress{Level 8, Doug McDonell Building}
  \postcode{3010}
  \city{Melbourne}
  \country{Australia}
}


\renewcommand{\shortauthors}{Messina and Pino, et al.}

\begin{abstract}
Every year physicians face an increasing demand of image-based diagnosis from patients, a problem that can be addressed with recent artificial intelligence methods. In this context, we survey works in the area of automatic report generation from medical images, with emphasis on methods using deep neural networks, with respect to: (1) Datasets, (2) Architecture Design, (3) Explainability and (4) Evaluation Metrics. Our survey identifies interesting developments, but also remaining challenges. Among them, the current evaluation of generated reports is especially weak, since it mostly relies on traditional Natural Language Processing (NLP) metrics, which do not accurately capture medical correctness.

\end{abstract}



\maketitle

\section{Introduction}

    The rapid and successful development of deep learning in research fields such as Computer Vision \cite{khan2019survey} and Natural Language Processing (NLP) \cite{otter2020survey} has found an important application area in healthcare, sustaining the promise of a future with more efficient and affordable medical care. 
    Research over the last five years shows a clear improvement in computer-aided detection (CAD), specifically in disease prediction from medical images \cite{wang2016discrimination, rajpurkar2017chexnet, gale2017detecting, tsai2019machine, hwang2019chestxray} as well as from Electronic Health Records (EHR) \cite{shickel2017deep}, by using deep neural networks (DNN) and treating the problem as supervised classification or segmentation tasks.
    Recently, Topol \cite{topol2019deep} indicates that the need for diagnosis and reporting from image-based examinations far exceeds the current medical capacity of physicians in the US. This situation promotes the development of automatic image-based diagnosis as well as automatic reporting.
    Furthermore, the lack of specialist physicians is even more critical in resource-limited countries \cite{rosman2019developing},
    and therefore the expected impacts of this technology would become even more relevant.

    However, the elaboration of high-quality medical reports from medical images, such as chest X-rays, computed tomography (CT) or magnetic resonance (MRI) scans, is a task that requires a trained radiologist with years of experience. 
    In this context, deep learning (DL) combined with other Artificial Intelligence (AI) techniques appears as a viable and promising solution to alleviate the physician scarcity problem, by both automating the report generation process and enhancing radiologists' performance through assisted report-generation. AI is set to have a significant impact on the medical imaging market and, hence, how radiologists work, with the ultimate goal of better patient outcomes.
    The pace of research in this area is rapid, and to the best of our knowledge, previous surveys on this topic \cite{pavlopoulos-etal-2019-survey, 10.1145/3286606.3286863, monshi2020survey} 
    do not cover aspects of explainability \cite{gunning2017explainable}, medical correctness and physician-centered evaluation. This article enhances these previous surveys by analyzing more than twenty additional works and datasets. 
    Furthermore, unlike previous surveys, in this article we pay special attention to explainable AI (XAI). XAI is a set of methods and technologies, which will allow physicians to better understand the rationale behind automatic reports from black-box algorithms \cite{guidotti2018survey}, potentially increasing trust for their actual clinical use. 

    \textbf{Contribution}. We summarize the state of research
    in automatic report generation from medical images.
    We perform an exhaustive review of the literature, consisting of 40 articles published in journals, conferences, and conference workshops proceedings.
    We first present an overview of the task (section \ref{section:task-overview}), followed by the survey methodology for search and selection of papers (section \ref{section:search-methodology}), and the research questions driving this research (section \ref{section:research-questions}).
    We then analyze papers regarding four dimensions: Datasets used (image modalities and clinical conditions, in section \ref{section:datasets}),
    Model Design (standard practices, input and output, visual and language components, domain knowledge, auxiliary tasks, and optimization strategies, in section \ref{section:model-design}),
    Explainability (section \ref{section:xai})
    and Evaluation Metrics (section \ref{section:metrics}).
    We also compare model performance of several articles (section \ref{section:comparison-papers}), identifying unsolved challenges across all reviewed papers and proposing potential avenues for future research (section \ref{section:challenges}).
    Lastly, we discuss the limitations of this work (section \ref{section:limitations}) and offer the main conclusions (section \ref{section:conclusion}).
    Our survey provides valuable insights to guide future research on automatic report generation from medical images.

\section{Task Overview}
\label{section:task-overview}

\begin{figure}[!t]
    \begin{tabular}{ll}
    \begin{minipage}{.36\textwidth}
        \includegraphics[width=0.52\linewidth]{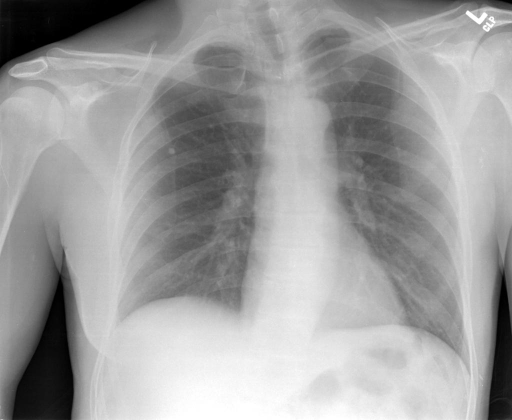}
        \includegraphics[width=0.35\linewidth]{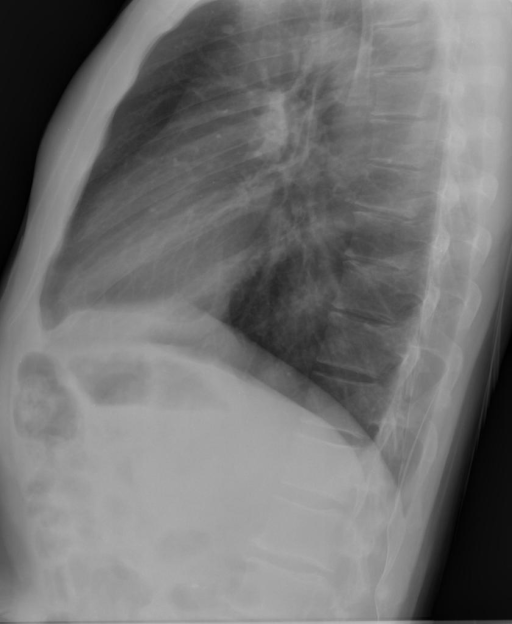}
        \\
        \raggedright
        \textbf{Manual tags:} Calcified Granuloma/lung/upper lobe/right\\
        \textbf{Automatic tags:} Calcified granuloma
    \end{minipage}
    &
    \begin{minipage}{.60\textwidth}
        \raggedright 
        \textbf{Comparison:}
        Chest radiographs XXXX.\\
        \textbf{Indication:}
        XXXX-year-old male, chest pain.\\
        \textbf{Findings:}
        The cardiomediastinal silhouette is within normal limits for size and contour. The lungs are normally inflated without evidence of focal airspace disease, pleural effusion, or pneumothorax. Stable calcified granuloma within the right upper lung. No acute bone abnormality.\\
        \textbf{Impression:}
        No acute cardiopulmonary process. \\
    \end{minipage}
    \\
    \end{tabular}
    \caption{Example from the IU X-ray dataset, frontal and lateral chest x-rays from a patient, alongside the natural language report and the annotated tags. XXXX is used for anonimization of the report.}
    \Description{Frontal and lateral chest x-rays, manually and automatically annotated tags, and a written report with four sections.}
    \label{figure:iu-xray-example}
\end{figure}

From a purely computational perspective, the following is the main task addressed by most articles analyzed in this survey: given as input one or more medical images of a patient, a text report is output that is as similar as possible to one generated by a radiologist.
From a machine learning point of view, creating a system that performs such a task would require learning a \textit{generative model} from instances of reports written by radiologists. 
Figure \ref{figure:iu-xray-example} presents one example of such a report, taken from the IU X-ray dataset \cite{10.1093/jamia/ocv080}.
We see two input X-ray images (frontal and lateral), and below them some annotations (Tags) --some manually annotated by a radiologist and others automatically annotated--, and on the right side the report with four different sections (comparison, indication, findings, and impression). 
If we consider the clinical workflow of generating a medical imaging report, several aspects should be taken into account before diving into a concrete implementation.

The first aspect is considering additional patient information in the process of report generation.
Most of the time, the physician asking for medical imaging is the primary care physician or a medical specialist. 
This implies that when radiologists write a report, they generally have patient-relevant clinical information, usually provided in the section \textit{Indication} as shown in Figure \ref{figure:iu-xray-example}. 
Also, the \textit{Comparison} section can provide information of a serial follow-up procedure, to evaluate the evolution of a patient over time (e.g., aneurysm, congenital heart disease).
Then, one decision can be whether or not to use these \textit{Indication} and \textit{Comparison} data to generate the sections \textit{Findings}, \textit{Impression}, or both of them.

Second, the model for report generation should consider the diversity on medical images as well as body regions and conditions. There are several types of medical images, such as X-rays, CT, MRI, PET and SPECT. This implies that a model for text report generation that deals with only one type of input medical image might not solve it for other types. Also, ideally, a model should be able to generate reports from different parts of the human anatomy and diverse medical conditions. To adequately achieve this task, different body regions must have a balanced and sizable training set. Many works surveyed in this article focus on one specific part of the body and particular illnesses 
which limits the applicability of these methods to generalize to all possible diagnosis tasks. 

Lastly, even if an AI system has perfect 
report generation accuracy, we might wonder if we can trust a machine in such a critical domain. One of the reasons for preferring a radiologist rather than an automated, highly accurate AI system 
is the chance of understanding the rationale behind the findings and impressions. In this sense, explainable AI \cite{gunning2017explainable} is of great importance 
in securing their adoption in a clinical setting.

\section{Survey Methodology: Search and selection of papers}
\label{section:search-methodology}

To collect the papers reviewed, we performed three main steps: retrieval, selection, and exclusion.
We further describe each step in the following paragraphs.

\textbf{Study retrieval}. To retrieve the articles we used seven search engines, namely Google Scholar, PubMed, Scopus, ACM Digital Library, Web of Knowledge, IEEE Xplore and Springer; and \surveyUpdate{two specific queries, plus other more relaxed queries}, described in Table \ref{table:methodology-amounts}.
\surveyUpdate{The relaxed queries returned articles already found with the two main queries.}
In this step we only considered journals, conference and conference workshop proceedings.

\textbf{Study selection}. Given the query results, a selection was performed applying inclusion criteria by reading title, abstract, and keywords of each paper.
If there was uncertainty after reading these sections, we included it for revision and decided afterward if it should be excluded with exclusion criteria.
The inclusion criteria were the following:
at least a part of the study focused on report generation from medical images.
The images can be from any kind (e.g., X-ray, MRI scans, CT scans), must be from humans, and may include one or more pathologies of any type\footnote{In practice, most datasets reviewed present one or more pathologies, since the detection of medical conditions is one of the main motivations of these studies.}.
The report must be in natural language form, comprising at least one or more sentences, and must be automatically or semi-automatically generated by a computational system that employs a DL technique.
Note that the method may contain steps that do not involve DL, such as rule-based decisions.
The system must receive as input one or more medical images, and it also might receive additional input, such as patient clinical history.
A semi-automated system may include a human in the process, expressly, by using additional input provided by the human.
We included 45 works in total.

\textbf{Study exclusion}. After thoroughly reading each paper selected, we used two exclusion criteria to discard works that were not relevant for this survey.
First, if the paper did not propose a specific computational approach to solve the report generation problem, for example, if presented a web application using existing methods, or presented an assessment of feasibility.
Second, if the task being addressed was different from natural language report generation from medical images, for example, report summarizing, disease classification from images, medical image segmentation, or any others.
We ruled out 5 works with these exclusion criteria, leaving a total of 40 papers.
The amount of papers found in each step is detailed in Table \ref{table:methodology-amounts}.




\vspace{-3mm}
\begin{table}[htbp]
    \begin{center}
        \begin{tabular}{|c|c|c|c|c|c|c|c|c|}
            \hline
            Query & Google Scholar & PubMed & Scopus & ACM & WoK & IEEE Xplore & Springer & Total \\
            \hline
            1 & 32 & 1 & 19 & 2 & 9 & 7 & 13 & 34 \\
            2 & 21 & 2 & 20 & 2 & 11 & 3 & 18 & 37 \\
            \hline\hline
            \multicolumn{8}{|l|}{\raggedright{Selected with inclusion criteria (all queries)}} & 45 \\
            \hline
            \multicolumn{8}{|l|}{\raggedright{Discarded with exclusion criteria \cite{10.1145/3204949.3208113, 10.1007/978-3-030-37446-4_10, 10.3389/fneur.2019.00007, 8588468, Loveymi2019}}} & 5 \\
            \hline
            \multicolumn{8}{    |p{0.9\linewidth}|}{\raggedright{\textbf{Total articles} \cite{jing2017automatic, liu2019clinically, huang2019multi, yuan2019automatic, li2018hybrid, li2019knowledge, wang2018tienet, xue2018multimodal, zhang2017mdnet, li2019vispi, xiong2019reinforced, singh2019chest, maksoud2019coral8, gale2018producing, 10.1007/978-3-030-00934-2_78, 8935910, 8970668, 10.1007/978-3-030-33850-3_8, 10.1007/978-3-030-04224-0_24, 10.1007/978-3-030-32251-9_37, gasimova2019automated, 10.1007/978-981-15-4015-8_15, harzig2019addressing, biswal2020clara, 10.1007/978-3-030-18590-9_64, Zeng2018, 10.1145/3343031.3356066, 10.1007/978-3-030-20351-1_10, 10.1007/978-3-030-26763-6_66, zhang2020radiology, 10.1007/978-3-030-00937-3_22, 10.1145/3357254.3357256, jing2019show, Shin_2016_CVPR, 10.1007/978-3-319-98932-7_21, Shin_2016_CVPR, zhang2017mdnet, Zeng2018, kisilev2016medical, moradi2016cross, wu2017generative, spinks2019justifying, zeng2020deep}}} & \textbf{40} \\
            \hline
        \end{tabular}
        \caption{Papers found for each query and database, and included or discarded with different criteria.
        WoK stands for Web of Knowledge.
        In both queries, only papers from journal, conference or conference workshops proceedings were included.\\
        Query 1: (medical OR medicine OR health) AND "report generation" AND (images OR image).\\
        Query 2: (medical OR medicine OR health) AND (images OR image) AND (report OR diagnostic OR description OR caption) AND (generation OR automatic) in ABSTRACT.\\
        \surveyUpdate{
        Relaxed queries: (medical report generation), (medical report image), (diagnostic captioning).
        }
        }
        \label{table:methodology-amounts}
    \end{center}
    \vspace{-7mm}
\end{table}

\section{Research Questions}
\label{section:research-questions}

This survey aims to answer the following research questions regarding the task of \textit{natural language report generation from medical images}:

\begin{enumerate}
    \item What datasets are used in this area?
    What diseases and imaging techniques are considered?

    \item
    What deep learning methods are the most commonly employed?
    
    \item
    What explainability or interpretability techniques are used?
    
    \item How are the proposed models evaluated?
    What metrics are used?
    
    \item How is the performance of the automatic methods?
    Which method can be considered \textit{state of the art} or showing the best performance?
    
    \item What are the main unsolved challenges? What are the potential avenues for future work?
\end{enumerate}

\section{Analysis of papers reviewed}

\subsection{Datasets}
\label{section:datasets}

We identify 18 \textit{report datasets} containing images and reports written by experts,
and 9 \textit{classification datasets}, which provide an image and the presence or absence of a list of abnormalities.
Most of the collections are publicly available (10 and 8 report and classification datasets, respectively), while the rest are proprietary.
In most cases, the datasets focus on one or more pathologies, and include both samples with presence and absence of these.
Table \ref{table:datasets-amounts} presents the main characteristics for the public collections, including a list of papers that used them. 
We next discuss the main remarks regarding report and classification datasets. 

\newcommand{\datasetSpanish}{$^{\textrm{(sp)}}$}
\newcommand{\datasetPortuguese}{$^{\textrm{(pt)}}$}
\newcommand{\datasetChinese}{$^{\textrm{(ch)}}$}
\newcommand{\datasetPendingRelease}{$^{\textrm{(1)}}$}
\newcommand{\datasetPMC}{$^{\textrm{(2)}}$}
\newcommand{\datasetROCO}{$^{\textrm{(3)}}$}
\newcommand{\datasetVideoFrames}{$^{\textrm{(4)}}$}
\newcommand{\datasetNone}{$^{\textrm{(5)}}$}

\begin{table}[htbp]
\scalebox{0.78}{
\begin{tabular}{|p{0.31\linewidth}|c|p{0.24\linewidth}|c|c|c|p{0.18\linewidth}|}
\hline
\textbf{Dataset} & \textbf{Year} & \textbf{Image Type} & \textbf{\# images} & \textbf{\# reports} & \textbf{\# patients} & \textbf{Used by papers} \\ \hline
\multicolumn{7}{|c|}{\textbf{Report datasets}} \\ \hline
IU X-ray \cite{10.1093/jamia/ocv080} & 2015 & Chest X-Ray & 7,470 & 3,955 & 3,955 & \cite{jing2017automatic, liu2019clinically, huang2019multi, yuan2019automatic, li2018hybrid, li2019knowledge, wang2018tienet, xue2018multimodal, li2019vispi, xiong2019reinforced, singh2019chest, 8970668, 10.1007/978-3-030-33850-3_8, gasimova2019automated, 10.1007/978-981-15-4015-8_15, harzig2019addressing, biswal2020clara, 10.1007/978-3-030-18590-9_64, 10.1007/978-3-030-20351-1_10, zhang2020radiology, jing2019show, Shin_2016_CVPR} \\ \hline
MIMIC-CXR \cite{johnson2019mimicv1, johnson2019mimiccxrjpg} & 2019 & Chest X-Ray & 377,110 & 227,827 & 65,379 & \cite{liu2019clinically} \\ \hline
PadChest\datasetSpanish \cite{bustos2019padchest} & 2019 & Chest X-Ray & 160,868 & 109,931 & 67,625 & None\datasetNone \\ \hline
\raggedright{ImageCLEF Caption 2017 \cite{eickhoff2017overview} } & 2017 & Biomedical\datasetPMC & 184,614 & 184,614 & - & \cite{10.1007/978-3-319-98932-7_21} \\ \hline
\raggedright{ImageCLEF Caption 2018 \cite{de2018overview} } & 2018 & Biomedical\datasetPMC & 232,305 & 232,305 & - & None\datasetNone \\ \hline 
ROCO \cite{dataset2018roco} & 2018 & Multiple radiology\datasetROCO & 81,825 & 81,825 & - & None\datasetNone \\ \hline
PEIR Gross \cite{jing2017automatic} & 2017 & Gross lesions & 7,442 & 7,442 & - & \cite{jing2017automatic} \\ \hline
INBreast\datasetPortuguese \cite{moreira2012inbreast} & 2012 & Mammography X-ray & 410 & 115 & 115 & \cite{10.1007/978-3-030-26763-6_66, 10.1145/3357254.3357256} \\ \hline
STARE \cite{hoover1975stare} & 1975 & Retinal fundus & 400 & 400 & - & None\datasetNone \\ \hline 
RDIF\datasetPendingRelease \cite{maksoud2019coral8} & 2019 & Kidney Biopsy & 1,152 & 144 & 144 & \cite{maksoud2019coral8} \\ \hline \hline
\multicolumn{7}{|c|}{\textbf{Classification datasets}} \\ \hline
CheXpert \cite{irvin2019chexpert} & 2019 & Chest X-Ray & 224,316 & 0 & 65,240 & \cite{yuan2019automatic, zhang2020radiology} \\ \hline
ChestX-ray14 \cite{Wang_2017_CVPR} & 2017 & Chest X-Ray & 112,120 & 0 & 30,805 & \cite{li2019knowledge, wang2018tienet, li2019vispi, xiong2019reinforced, biswal2020clara, 10.1007/978-3-030-20351-1_10, jing2019show} \\ \hline
LiTS \cite{christ2017lits} & 2017 & Liver CT scans & 200 & 0 & - & \cite{10.1007/978-3-030-00934-2_78} \\ \hline 
\raggedright{ACM Biomedia 2019 \cite{10.1145/3343031.3356058}} & 2019 & Gastrointestinal tract \datasetVideoFrames & 14,033 & 0 & - & \cite{10.1145/3343031.3356066} \\ \hline
DIARETDB0 \cite{kauppi2006diaretdb0} & 2006 & Retinal fundus & 130 & 0 & - & \cite{wu2017generative} \\ \hline
DIARETDB1 \cite{kalviainen2007diaretdb1} & 2007 & Retinal fundus & 89 & 0 & - & \cite{wu2017generative} \\ \hline
Messidor \cite{decenciere2014feedback, abramoff2013automated} & 2013 & Retinal fundus & 1,748 & 0 & 874 & \cite{wu2017generative} \\ \hline
DDSM \cite{heath2001digital} & 2001 & Mammography X-ray & 10,480 & 0 & - & \cite{kisilev2016medical} \\ \hline
\end{tabular}
}
\caption{Public datasets used in the literature.
    All reports are written in English, except those marked with \datasetSpanish{} which are in Spanish, and \datasetPortuguese{} in Portuguese.
    Other notes,
    \datasetPendingRelease: the RDIF dataset is pending release.
    \datasetPMC: for the ImageCLEF datasets, images were extracted from PubMed Central papers and filtered automatically in order to keep only clinical images, but some unintended samples from other domains are also included.
    \datasetROCO: contains multiple modalities, namely CT, Ultrasound, X-Ray, Fluoroscopy, PET, Mammography, MRI, Angiography and PET-CT.
    \datasetVideoFrames: the images are frames extracted from videos.
    \datasetNone: none of the papers reviewed used this dataset.
    }
\label{table:datasets-amounts}
\vspace{-6mm}
\end{table}

The third column in Table \ref{table:datasets-amounts} lists the image modalities for each dataset, showing chest X-rays concentrates most of the efforts in report datasets \cite{10.1093/jamia/ocv080, johnson2019mimiccxrjpg, bustos2019padchest, li2018hybrid, 8935910}, though there are also datasets with biomedical images from varied types \cite{eickhoff2017overview, de2018overview, dataset2018roco, jing2017automatic}, mammography \cite{moreira2012inbreast} and hip X-rays \cite{gale2017detecting}, ultrasound images \cite{10.1007/978-3-030-32251-9_37, Zeng2018}, retinal images \cite{hoover1975stare}, doppler echocardiographies \cite{moradi2016cross}, cervical images \cite{10.1007/978-3-030-04224-0_24}, and kidney \cite{maksoud2019coral8} and bladder biopsies \cite{zhang2017mdnet}.
This adds an extra challenge, since different kinds of exams may need different solutions, as the clinical conditions will be diverse.
For example, a fundus retinal image may differ significantly from a chest X-ray; or a radiologist analyzing an X-ray may follow a different procedure than a pathologist reading a biopsy.

From the public report datasets, IU X-ray \cite{10.1093/jamia/ocv080} is the most commonly used, consisting of 7,470 frontal and lateral chest X-rays and 3,955 reports.
Additionally, each report was manually annotated with Medical Subject Heading (MeSH)\footnote{\url{https://www.nlm.nih.gov/mesh/meshhome.html}} \cite{rogers1963medical} and RadLex \cite{langlotz2006radlex} terms, and automatically annotated with MeSH terms using the MTI \cite{mork2013nlm} system plus the negation tool from MetaMap \cite{aronson2010overview}.
Figure \ref{figure:iu-xray-example} shows a sample image and report from this dataset.
Note that for deep learning methods, the amount of data may seem insufficient, compared to general domain datasets with millions of samples, such as ImageNet \cite{deng2009imagenet}.
This issue could be addressed with pre-training or data augmentation techniques.
Also, this may be partially solved with the more recent datasets MIMIC-CXR \cite{johnson2019mimiccxrjpg} or PadChest \cite{bustos2019padchest}, which contain 377,110 and 160,868 images respectively, but have not been widely used yet.

All report datasets include images and reports, and most of them also include labels for each report.
Furthermore, INbreast \cite{moreira2012inbreast} includes contours locating the labels in the images,
the Ultrasound collection \cite{Zeng2018, zeng2020deep} includes bounding boxes locating organs,
and IU X-ray \cite{10.1093/jamia/ocv080} and RDIF \cite{maksoud2019coral8} include additional text written by the physician who requested the exam. 
The complete detail of additional information is shown in Table \ref{table:datasets-extra-data-full} in appendix \ref{section:appendix:datasets}.
This information can be leveraged as a supplementary context to further improve the system performance. On the one hand, the labels and image localization can be used to design auxiliary tasks (see section \ref{section:aux-tasks}), and to further evaluate the text generation process (see section \ref{section:metrics}).
On the other hand, the indication may contain additional information not present in the image, such as a patient's previous condition, which in some cases may be essential to address the task \cite{maksoud2019coral8}.

Lastly, many works use classification datasets, which do not provide a report for each image, but a set of clinical conditions or abnormalities present or absent in the image.
In most cases, this kind of information is used to perform image classification as pre-training, an intermediate, or an auxiliary task to generate the report.
One remarkable case is the CheXpert dataset \cite{irvin2019chexpert}, which contains 224,316 images, and was also presented with the CheXpert labeler, an automatic rule-based tool that annotates 14 labels (abnormalities) as present, absent or uncertain from the natural language reports.
This tool was used to label the images from the dataset, is also used in MIMIC-CXR \cite{johnson2019mimiccxrjpg} to tag the reports, and in some works to evaluate the generated reports, as discussed in the Metrics section (\ref{section:metrics}).
Notice the classification dataset list is not comprehensive, as it only includes datasets that were used in at least one of the reviewed works.

\subsubsection*{Synthesis}
The datasets cover multiple image modalities and body parts, though most efforts focus on chest X-rays.
This opens a potential research avenue to explore other image types and diseases, using existing solutions or raising new methods.
Additionally, most collections provide valuable supplementary information, such as abnormality tags and/or localization, which can be used to design auxiliary tasks and to evaluate the performance.

\subsection{Model Design}
\label{section:model-design}

\begin{figure}
    \centering
    \includegraphics[width=\textwidth]{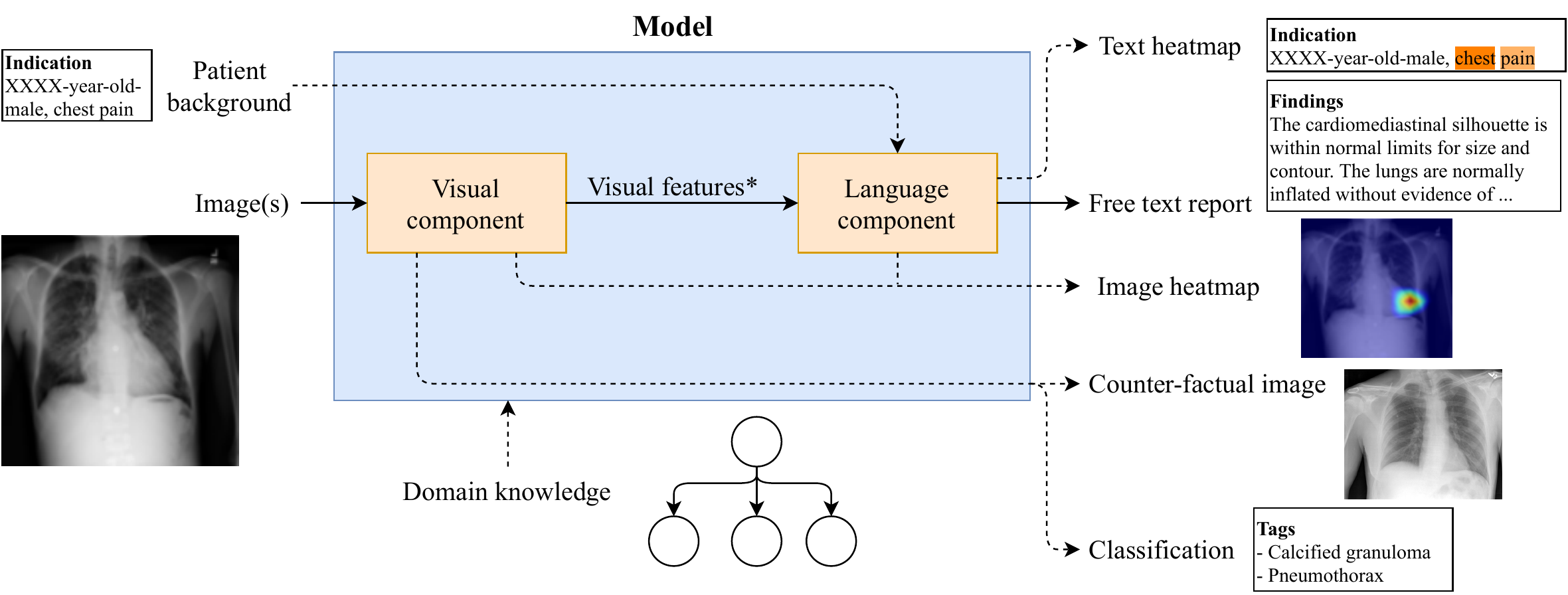}
    \caption{General scheme of components of the architectures reviewed, including inputs on the left and outputs on the right. The blue box represents the whole model, while the orange boxes show the inner components.
    Solid line arrows show the flow shared by almost every work reviewed, while dashed line arrows show optional inputs and outputs seen only in some papers.
    Note *: in some works, the visual component may transfer classification or segmentation outputs besides or instead of \textit{visual features}.
    }
    \Description{General model architecture found in the literature review. The model consists of a visual and a language component. There are two possible inputs, image(s) and patient background; and five possible outputs, free text report, a heatmap over the input text, a heatmap over the input image, a counter-factual image, and classification output. Additionally, explicit domain knowledge can be incorporated in the model.}
    \label{figure:architecture-scheme}
\vspace{-6mm}    
\end{figure}

In this section, we present an analysis of existent DL model designs, starting with a general overview of common design practices.
Most models in the literature follow a standard design pattern. There is a visual component consisting at its core of a Convolutional Neural Network (CNN) \cite{krizhevsky2012imagenet} that processes one or more input images in order to extract visual features.
Then, a language component follows, typically based on well-known NLP neural architectures (e.g., LSTM \cite{hochreiter1997long}, BiLSTM \cite{graves2005framewise}, GRU \cite{chung2014empirical}, Transformer \cite{vaswani2017attention}) responsible for text processing and report generation.
Also, a widespread practice for the language component is to retrieve the visual information in an adaptive manner via an attention mechanism, as the report is written. Many papers follow variations of this pattern inspired by influential works from the image captioning domain \cite{vinyals2015show, xu2015show}, which are frequently cited and used as baselines.
Optionally, some models receive or generate additional input or output, and a few models incorporate some form of domain knowledge explicitly in the generation process.
Figure \ref{figure:architecture-scheme} presents a summary illustration of a general model architecture found in the literature.
Next, we analyze model designs according to 6 dimensions: (1) input and output, (2) visual component, (3) language component, (4) domain knowledge, (5) auxiliary tasks and (6) optimization strategies.

\subsubsection{Input and Output}
\label{section:input-output}


\begin{table}[htbp]
\small
\scalebox{0.78}{
\begin{tabular}{|p{0.18\linewidth}|p{0.45\linewidth}|p{0.5\linewidth}|}
\hline
\textbf{Category} & \textbf{Value or Type} & \textbf{Used by papers} \\ \hline
\multicolumn{3}{|c|}{\textbf{Input}} \\ \hline
\multirow[t]{14}{\linewidth}{Image Type} & Chest X-Ray & \cite{jing2017automatic, liu2019clinically, huang2019multi, yuan2019automatic, li2018hybrid, li2019knowledge, wang2018tienet, xue2018multimodal, li2019vispi, xiong2019reinforced, singh2019chest, 8935910, 8970668, 10.1007/978-3-030-33850-3_8, gasimova2019automated, 10.1007/978-981-15-4015-8_15, harzig2019addressing, biswal2020clara, 10.1007/978-3-030-18590-9_64, 10.1007/978-3-030-20351-1_10, zhang2020radiology, jing2019show, Shin_2016_CVPR, spinks2019justifying} \\
& Mammography X-ray & \cite{10.1007/978-3-030-26763-6_66, 10.1145/3357254.3357256, kisilev2016medical} \\
& Hip X-Ray & \cite{gale2018producing} \\
& Ultrasound video frames & \cite{10.1007/978-3-030-32251-9_37, Zeng2018, kisilev2016medical, zeng2020deep} \\
& CW Doppler echocardiography & \cite{moradi2016cross} \\
& Gastrointestinal tract examination frames & \cite{10.1145/3343031.3356066} \\
& Gross lesions & \cite{jing2017automatic} \\
& Bladder biopsy & \cite{zhang2017mdnet} \\
& Kidney biopsy & \cite{maksoud2019coral8} \\
& Liver tumor CT scans & \cite{10.1007/978-3-030-00934-2_78} \\
& Cervical neoplasm WSI & \cite{10.1007/978-3-030-04224-0_24} \\
& Spine MRI & \cite{10.1007/978-3-030-00937-3_22} \\
& Fundus retinal images & \cite{wu2017generative}\\
& Biomedical images & \cite{10.1007/978-3-319-98932-7_21} \\
\hline
\multirow[t]{3}{\linewidth}{Number of images} & 1 & \cite{jing2017automatic, liu2019clinically, huang2019multi, wang2018tienet, zhang2017mdnet, li2019vispi, xiong2019reinforced, singh2019chest, gale2018producing, 8935910, 8970668, 10.1007/978-3-030-33850-3_8, 10.1007/978-3-030-04224-0_24, 10.1007/978-3-030-32251-9_37, gasimova2019automated, 10.1007/978-981-15-4015-8_15, harzig2019addressing, biswal2020clara, Zeng2018, 10.1145/3343031.3356066, 10.1007/978-3-030-20351-1_10, 10.1007/978-3-030-26763-6_66, 10.1007/978-3-030-00937-3_22, 10.1145/3357254.3357256, jing2019show, Shin_2016_CVPR, 10.1007/978-3-319-98932-7_21, kisilev2016medical, moradi2016cross, wu2017generative, spinks2019justifying, zeng2020deep} \\
& 2 & \cite{yuan2019automatic, li2018hybrid, li2019knowledge, xue2018multimodal, 10.1007/978-3-030-18590-9_64, zhang2020radiology} \\
& Any & \cite{maksoud2019coral8, 10.1007/978-3-030-00934-2_78} \\
\hline
\multirow[t]{4}{\linewidth}{Text} & Indication & \cite{huang2019multi, maksoud2019coral8} \\
& Indication and findings & \cite{10.1007/978-3-030-33850-3_8} \\
& Prefix sentence and keywords & \cite{biswal2020clara} \\
& Partial report or caption & \cite{10.1007/978-3-030-32251-9_37, 10.1007/978-981-15-4015-8_15} \\
\hline
\multicolumn{3}{|c|}{\textbf{Output}} \\ \hline
\multirow[t]{5}{\linewidth}{Report} & Generative multi-sentence (unstructured) & \cite{jing2017automatic, liu2019clinically, huang2019multi, yuan2019automatic, wang2018tienet, xue2018multimodal, li2019vispi, xiong2019reinforced, singh2019chest, maksoud2019coral8, 8935910, 8970668, 10.1007/978-3-030-33850-3_8, 10.1007/978-981-15-4015-8_15, harzig2019addressing, 10.1007/978-3-030-18590-9_64, 10.1007/978-3-030-20351-1_10, 10.1007/978-3-030-26763-6_66, zhang2020radiology, jing2019show} \\
& Generative multi-sentence structured & \cite{zhang2017mdnet, 10.1007/978-3-030-00934-2_78} \\
& Generative single-sentence & \cite{gale2018producing, 10.1007/978-3-030-32251-9_37, gasimova2019automated, Zeng2018, 10.1145/3357254.3357256, Shin_2016_CVPR, 10.1007/978-3-319-98932-7_21, wu2017generative, spinks2019justifying, zeng2020deep} \\
& Template-based & \cite{10.1007/978-3-030-04224-0_24, 10.1145/3343031.3356066, 10.1007/978-3-030-00937-3_22, kisilev2016medical, moradi2016cross} \\
& Hybrid template - generation/edition & \cite{li2018hybrid, li2019knowledge, biswal2020clara} \\
\hline
\multirow[t]{6}{\linewidth}{Classification} & MeSH concepts or similar & \cite{jing2017automatic, yuan2019automatic, 8935910, 8970668, 10.1007/978-3-030-33850-3_8, harzig2019addressing, 10.1145/3343031.3356066, 10.1007/978-3-030-26763-6_66, Shin_2016_CVPR} \\
& Abnormalities/diseases presence or absence & \cite{li2019knowledge, wang2018tienet, li2019vispi, xiong2019reinforced, biswal2020clara, Zeng2018, zhang2020radiology, jing2019show, spinks2019justifying, zeng2020deep} \\
& Abnormalities/diseases characterization or severity level & \cite{zhang2017mdnet, gale2018producing, 10.1007/978-3-030-04224-0_24, kisilev2016medical} \\
& Body parts or organs & \cite{10.1007/978-3-030-32251-9_37, Zeng2018, moradi2016cross, zeng2020deep} \\
& Image modality & \cite{10.1007/978-3-319-98932-7_21} \\
& Normal or abnormal sentence & \cite{harzig2019addressing, 10.1007/978-3-030-18590-9_64, jing2019show} \\
\hline
\multirow[t]{9}{\linewidth}{Image Heatmap} & Attention-based per word & \cite{liu2019clinically, wang2018tienet, zhang2017mdnet} \\
& Attention-based per sentence & \cite{jing2017automatic, huang2019multi, xue2018multimodal, 10.1007/978-3-030-20351-1_10} \\
& Attention-based per report & \cite{li2019knowledge} \\
& CAM \cite{zhou2016learning} & \cite{10.1007/978-3-030-04224-0_24, 10.1145/3343031.3356066} \\
& Grad-CAM \cite{selvaraju2016grad} & \cite{yuan2019automatic, li2019vispi} \\
& SmoothGrad \cite{smilkov2017smoothgrad} & \cite{gale2018producing} \\
& Activation-based attention \cite{zagoruyko2016paying} & \cite{spinks2019justifying} \\
& Bounding Box (Faster R-CNN \cite{ren2015faster}) & \cite{zeng2020deep, kisilev2016medical} \\
& Disease and body part pixel-level classification & \cite{10.1007/978-3-030-00934-2_78, 10.1007/978-3-030-00937-3_22} \\
\hline
\multirow[t]{1}{\linewidth}{Text Heatmap} & Attention based per word & \cite{huang2019multi} \\
\hline
\multirow[t]{1}{\linewidth}{Others} & Counter-factual example generation & \cite{spinks2019justifying} \\
\hline
\end{tabular}
}
\caption{Summary of input and output analysis of the reviewed literature.}
\label{table:io}
\vspace{-10mm}
\end{table}

Table \ref{table:io} presents a summary of this analysis.

\textit{Input}.
With respect to image type, most papers (24) used chest X-rays, whereas the other papers are more or less equally distributed over other image types.
A total of 32 models receive a single image (e.g. a single chest X-ray view), 6 models receive 2 images (both frontal and lateral chest X-ray views), and 2 models receive an arbitrary number of images (e.g. multi-slice abdominal CT scans).
Most models in the literature only handle visual input.
However, 6 works \cite{huang2019multi, maksoud2019coral8, 10.1007/978-3-030-33850-3_8, biswal2020clara, 10.1007/978-3-030-32251-9_37, 10.1007/978-981-15-4015-8_15} explored the use of complementary input text, reporting performance gains in most cases.
For example, two works \cite{huang2019multi, maksoud2019coral8} encode an \textit{indication} paragraph with a BiLSTM. Similarly, MTMA \cite{10.1007/978-3-030-33850-3_8} encodes the report's \textit{indication} and \textit{findings} sections with a BiLSTM per sentence first, and then a LSTM produces a final vector representation.
Similarly, two works \cite{10.1007/978-3-030-32251-9_37, 10.1007/978-981-15-4015-8_15} use LSTM/BiLSTM to encode a partial report or caption as input, in order to predict the next word.
Unlike other works, CLARA \cite{biswal2020clara} uses a software package, \textit{Lucene} \cite{milosavljevic2010retrieval}, to perform text-based retrieval of report templates. The input text is processed by \textit{Lucene} as a search query, and the retrieved templates are paraphrased by an encoder-decoder network to generate the final report.

\textit{Output}. All models output a natural language report. According to the extension of the report and the general strategy used to produce it, we group papers into five categories:
(1) \textit{Generative multi-sentence (unstructured)}: these models generate a multi-sentence report, word by word, with freedom to decide the number of sentences and the words in each sentence.
(2) \textit{Generative multi-sentence structured}: similar to the previous category, but always output a fixed number of sentences, and each sentence always has a pre-defined topic. These models are designed for datasets where reports follow a rigid structure.
(3) \textit{Generative single-sentence}: generate a report word by word, but only output a single sentence. These models are designed for datasets with simple one-sentence reports.
(4) \textit{Template-based}: use human-designed templates to produce the report, for example performing a classification task followed by if-then rules, template selection and template filling. This simplifies the report generation task for the model, at the expense of making it less flexible and requiring the human designing of templates and rules.
And lastly (5) \textit{Hybrid template - generation/edition}: use templates and also have the freedom to generate sentences word by word. This can be accomplished by choosing between a template or generating a sentence from scratch \cite{li2018hybrid}, or by editing/paraphrasing a previously selected template \cite{li2019knowledge, biswal2020clara}.

In addition to the report itself, many models also output complementary classification predictions, such as presence or absence of abnormalities or diseases, MeSH concepts, body parts or organs, among others.
These are often referred to as labels or tags, and are commonly used in the language component, as will be discussed in section \ref{section:language-component}.
%
Many models can also output heatmaps over an image highlighting relevant regions using different techniques, such as explicit visual attention weights computed during report generation, saliency maps methods (e.g., CAM, Grad-CAM, SmoothGrad or activation-based attention), bounding box regression, and pixel-level classification (image segmentation).
Also, one model \cite{huang2019multi} can output a heatmap over its input text and one model \cite{spinks2019justifying} can generate a counter-factual example to justify its decision. We will discuss all these outputs more in detail and their use in the explainability section (\ref{section:xai}).

\subsubsection{Visual Component}
\label{section:visual-component}

The most important observation is that all surveyed works use CNNs to process the input images. This is not surprising since CNNs have dominated the state of the art in computer vision for several years \cite{khan2019survey}.
The typical visual processing pipeline consists of a CNN that receives an input image and outputs a volume of feature maps of dimensions $W\times H\times C$, where $W$ and $H$ denote spatial dimensions (width and height) and $C$ denotes the channel dimensions (depth or number of feature maps).
These visual features are then leveraged by the language component to make decisions for report generation (e.g., which sentence to write, which template to retrieve, next word to output, etc.), typically by way of an attention mechanism.

However, some works did not strictly follow this pattern. For example, in two works \cite{8935910, 10.1007/978-3-030-26763-6_66} a CNN is used for multi-label classification of tags, which are then mapped to embedded vectors via embedding matrix lookup. Thus, the report generation module only has access to these tag vectors but no access to the visual features themselves.
Similarly, two works \cite{jing2017automatic, 8970668} classify and look up tag embedding vectors, but unlike the previous works, the language component uses co-attention to access both tags vectors and visual features simultaneously. Their ablation analysis showed that the semantic information provided by these tags complements the visual information and improves the model's performance in report generation.
Other works \cite{li2019knowledge, zhang2020radiology} used graph neural networks immediately after the CNN to encode the visual information in terms of medical concepts and their relations. Thus, the language component receives the intermediate graph representation instead of the raw visual features. The ablation analysis by Zhang et al. \cite{zhang2020radiology} showed some performance gains thanks to the graph neural network.
Vispi \cite{li2019vispi} implements a two-stage procedure, where two distinct CNNs are used. In the first stage a DenseNet 121 \cite{huang2017densely} classifies abnormalities in the image, and then Grad-CAM \cite{selvaraju2016grad} is used to localize and crop a region of the image for each detected class. Then, in the second stage the multiple image crops are treated as independent images and processed by a typical CNN+LSTM architecture, with ResNet 101 \cite{he2016deep} as the CNN. A similar idea was followed in RTMIC \cite{xiong2019reinforced}, where a DenseNet 121 is pretrained for classification in ChestX-ray14 \cite{Wang_2017_CVPR} and CAM is used to get image crops for each class.


\newcommand{\adhocnetwork}{$^{\textrm{(*)}}$}

\begin{table}[tbp]
\scalebox{0.8}{
\begin{tabular}{|l|l|}
\hline
 \textbf{Architecture}  & \textbf{Used by papers}
 \\ \hline
 DenseNet \cite{huang2017densely} & \cite{liu2019clinically, li2019vispi, xiong2019reinforced, zhang2020radiology, li2018hybrid, li2019knowledge, gale2018producing, 8970668, biswal2020clara}\\
 ResNet \cite{he2016deep} & \cite{huang2019multi, yuan2019automatic, xue2018multimodal, harzig2019addressing, 10.1007/978-3-030-20351-1_10, li2019vispi, 8935910, wang2018tienet, gasimova2019automated, jing2019show, 10.1007/978-3-030-04224-0_24}\\
 VGG \cite{simonyan2014very} & \cite{jing2017automatic, 10.1007/978-3-319-98932-7_21, maksoud2019coral8, 10.1007/978-3-030-32251-9_37, gasimova2019automated, 10.1007/978-981-15-4015-8_15, Zeng2018, 10.1145/3357254.3357256, kisilev2016medical, zeng2020deep, moradi2016cross}\\
 Faster R-CNN \cite{ren2015faster} & \cite{kisilev2016medical, zeng2020deep}\\
 Inception V3 \cite{szegedy2016rethinking} & \cite{singh2019chest}\\ 
 GoogLeNet \cite{szegedy2015going} & \cite{Shin_2016_CVPR}\\
 MobileNet V2 \cite{howard2017mobilenets} & \cite{10.1145/3343031.3356066}\\
 SRN \cite{zhu2017learning} & \cite{8935910}\\
 U-Net \cite{ronneberger2015u} & \cite{10.1007/978-3-030-26763-6_66}\\
 EcNet \adhocnetwork & \cite{zhang2017mdnet}\\
 FCN + shallow CNN \adhocnetwork & \cite{10.1007/978-3-030-00934-2_78}\\
 RGAN \adhocnetwork & \cite{10.1007/978-3-030-00937-3_22}\\
 StackGAN \cite{zhang2017stackgan} \textit{(slightly modified version)} \adhocnetwork & \cite{spinks2019justifying}\\
 CNN \adhocnetwork & \cite{10.1007/978-3-030-33850-3_8, spinks2019justifying}\\
 CNN \textit{(unspecified architecture)} &  \cite{10.1007/978-3-030-18590-9_64, wu2017generative}\\
 \hline

\end{tabular}
}
\caption{Summary of convolutional neural network architectures used in the literature. RGAN stands for recurrent generative adversarial network, FCN for fully convolutional network and EcNet is the name given in MDNet \cite{zhang2017mdnet} to the custom CNN used. \adhocnetwork: indicates an \textit{ad hoc} architecture.}
\label{tab:vc}
\vspace{-10mm}
\end{table}

We observe a wide variety of CNN architectures used in the literature, though most works employ standard designs. Table \ref{tab:vc} presents a summary. The most common ones are ResNet (11 works), VGG (11 works), and DenseNet (9 works). Other standard architectures used are Faster R-CNN, Inception V3, GoogLeNet, MobileNet V2, Spatial Regularization Network (SRN) and U-Net.
Five works used \textit{ad hoc} architectures not previously published (marked with (*) in Table \ref{tab:vc}).
For example, EcNet is an \textit{ad hoc} architecture used in MDNet \cite{zhang2017mdnet} and was proposed as an improvement over ResNet. However, the authors acknowledged that its design resembles DenseNet, which was published the same year (2017).
RGAN, proposed by Han et al. \cite{10.1007/978-3-030-00937-3_22}, is a novel architecture that follows the generative adversarial network (GAN) \cite{goodfellow2014generative} approach, with a generative module comprising the encoder and decoder parts of an atrous convolution autoencoder (ACAE) with a spatial LSTM between them.
Similarly, Spinks and Moens \cite{spinks2019justifying} used a slightly modified version of a StackGAN \cite{zhang2017stackgan} to learn the mapping from report encoding to chest X-ray images, and a custom CNN to learn the inverse mapping. Both are trained together, but only the latter is part of the report generation pipeline during inference.

\subsubsection{Language Component}
\label{section:language-component}


\begin{table}[tpb]
\scalebox{0.8}{
\begin{tabular}{|p{0.7\textwidth}|p{0.42\textwidth}|}
\hline
 \textbf{Architecture}  & \textbf{Used by papers}
 \\ \hline
 GRU &
 \cite{Shin_2016_CVPR}\\
 
 LSTM &
 \cite{singh2019chest, 8935910, gasimova2019automated, Zeng2018, 10.1007/978-3-030-26763-6_66, 10.1145/3357254.3357256, Shin_2016_CVPR, 10.1007/978-3-319-98932-7_21, wu2017generative, zeng2020deep}\\
 
 LSTM with attention &
 \cite{wang2018tienet, zhang2017mdnet, li2019vispi, gale2018producing, 10.1007/978-3-030-00934-2_78}\\
 
 Hierarchical LSTM with attention &
 \cite{jing2017automatic, liu2019clinically, huang2019multi, yuan2019automatic, zhang2020radiology, 8970668, 10.1007/978-3-030-33850-3_8}\\
 
 Hierarchical: Sentence LSTM + Dual Word LSTM (normal/abnormal) &
 \cite{harzig2019addressing, 10.1007/978-3-030-18590-9_64, jing2019show}\\
 
 Recurrent BiLSTM-attention-LSTM &
 \cite{xue2018multimodal, maksoud2019coral8, 10.1007/978-3-030-20351-1_10}\\
 
 Partial report encoding + FC layer (next word) &
 \cite{10.1007/978-3-030-32251-9_37, 10.1007/978-981-15-4015-8_15}\\ 
 
 Transformer &
 \cite{xiong2019reinforced}\\
 
 ARAE &
 \cite{spinks2019justifying}\\
 
 Template based &
 \cite{10.1007/978-3-030-04224-0_24, 10.1145/3343031.3356066, 10.1007/978-3-030-00937-3_22, kisilev2016medical, moradi2016cross}\\
 
 Hybrid template retrieval + generation/edition &
 \cite{li2019knowledge, biswal2020clara, li2018hybrid}\\
 \hline

\end{tabular}
}
\caption{Summary of language component architectures used in the literature. ARAE stands for adversarially regularized autoencoder.}
\label{tab:lc}
\end{table}

The job of the language component is to generate the report. In contrast to the visual component, in the literature we find a greater variety of approaches and creative ideas applied to this component. Table \ref{tab:lc} presents a high-level summary of this analysis.

The simplest approach is the use of a recurrent neural network, such as LSTM or GRU, to generate the full report word by word. Nine works \cite{singh2019chest, 8935910, gasimova2019automated, Zeng2018, 10.1007/978-3-030-26763-6_66, 10.1145/3357254.3357256, 10.1007/978-3-319-98932-7_21, wu2017generative, zeng2020deep} used LSTM and one work \cite{Shin_2016_CVPR} tried both GRU and LSTM. All these works have in common that the GRU/LSTM receives an encoding vector from the visual component at the beginning and the full report is decoded from it. This encoding vector is typically a vector of global features output by the CNN. However, two of these works \cite{8935910, 10.1007/978-3-030-26763-6_66} compute a weighted sum of tag embedding vectors and provide that as input to the LSTM.
Five works \cite{wang2018tienet, zhang2017mdnet, li2019vispi, gale2018producing, 10.1007/978-3-030-00934-2_78} used LSTM enhanced with an attention mechanism. In addition to the initial input, the LSTM equipped with attention can selectively attend to visual features from the visual component at each recurrent step. This typically leads to improved performance in all papers.

A known problem for recurrent networks such as LSTM is that they are not very good at generating very long texts \cite{pascanu2013difficulty}.
This is not a worrying issue when reports are short, 
however, it can become one for long multi-sentence reports.
Two papers \cite{zhang2017mdnet, 10.1007/978-3-030-00934-2_78} worked around this by generating each sentence independently with a single LSTM and then concatenating these sentences together. They accomplished this by providing the LSTM with a vector that indicates the sentence type as first input. This worked well in their case because the models were designed for structured reports, i.e., a fixed number of sentences per report and a fixed topic per sentence. Vispi \cite{li2019vispi} adopts a similar strategy: 
for each disease a dedicated LSTM generates the corresponding sentence, and the final report is the concatenation of them.

\begin{figure}
    \centering
    \includegraphics[width=\textwidth]{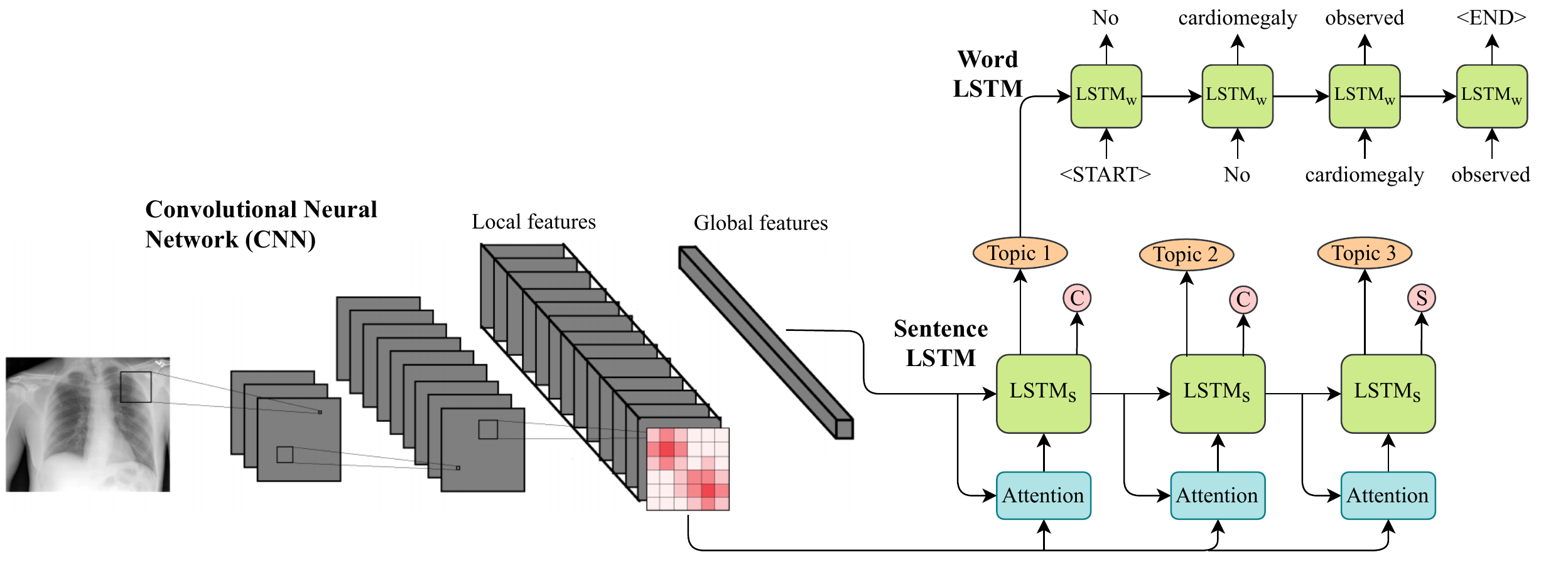}
    \caption{Illustration of a model following the \textit{Hierarchical LSTM with attention} approach, with attention at the sentence level. The visual component consists of a CNN. The \textit{global features} vector can be computed from the \textit{local features} in many ways, e.g. global average pooling.
    In each step the sentence LSTM generates a topic vector representing the current sentence, and decides whether to stop (S) generating or continue (C).
    }
    \label{figure:cnn+hlstm}
\vspace{-7mm}    
\end{figure}

To tackle the generation of unstructured multi-sentence reports, a group of papers followed what we call the \textit{Hierarchical LSTM with attention} approach: a Sentence LSTM generates a sequence of \textit{topic vectors}, and a Word LSTM receives a topic vector and generates a sentence word by word.
In this setting, the attention mechanism can be present at the sentence level, the word level or both.
Figure \ref{figure:cnn+hlstm} shows an illustrative example.
Seven works \cite{jing2017automatic, liu2019clinically, huang2019multi, yuan2019automatic, zhang2020radiology, 8970668, 10.1007/978-3-030-33850-3_8} followed this approach.
A common result in these papers is that a Hierarchical LSTM yields better performance in multi-sentence report generation than a single, flat LSTM.
A few papers \cite{harzig2019addressing, 10.1007/978-3-030-18590-9_64, jing2019show} went one step further and replaced the normal Word LSTM with a Dual Word LSTM: the model has a gating mechanism at the sentence level that decides if the sentence will describe an abnormality (e.g., a detected cardiomegaly) or a healthy case. Thus, there are two Word LSTMs, one for normal and one for abnormal sentences.
The goal is to improve the generation of abnormal sentences by having a Word LSTM that specializes in generating them. In contrast, a single Word LSTM for everything can lead to overlearning of normal sentences and underlearning of abnormal ones, as the latter are typically less frequent due to class imbalances in datasets. The ablation analyses of these works show performance gains, thanks to this approach.

Another approach for multi-sentence report generation is the \textit{Recurrent BiLSTM-attention-LSTM} approach. The basic idea is to have a LSTM generate one sentence at a time, each time conditioned on a BiLSTM based encoding of the previous sentence and the output of an attention mechanism. The process is repeated recurrently sentence by sentence until the full report is generated. Three papers used this approach \cite{maksoud2019coral8, xue2018multimodal, 10.1007/978-3-030-20351-1_10}.

Two works \cite{10.1007/978-3-030-32251-9_37, 10.1007/978-981-15-4015-8_15} approached report generation as simply learning to predict the next word given a partial report and an image. The models have dedicated components, such as LSTM and BiLSTM, for encoding the partial report and the image, and the next word is predicted by an FC layer.
This approach simplifies the task (i.e., predict the next word given everything that comes before), but in practice requires that the model be applied recurrently one word at a time to produce a full report, which has quadratic instead of linear complexity.

Only one work, RTMIC \cite{xiong2019reinforced}, has explored the use of the Transformer \cite{vaswani2017attention} architecture for report generation. In RTMIC multiple image crops are obtained using Grad-CAM, then from each crop a feature vector is obtained, and finally a Transformer converts these vectors into a report. The paper's results show some performance gains in CIDEr and BLEU with respect to some baselines that do not use the Transformer. Likewise, Spinks and Moens \cite{spinks2019justifying} were the only ones to use an adversarially regularized autoencoder (ARAE) \cite{zhao2017adversarially} to generate reports. Their model combines an ARAE with a StackGAN and a normal CNN, achieving better performance than a convolutional caption generation baseline in several NLP metrics.

We also identify a group of papers \cite{10.1007/978-3-030-04224-0_24, 10.1145/3343031.3356066, 10.1007/978-3-030-00937-3_22, kisilev2016medical, moradi2016cross} following a \textit{Template based} approach. The language component in these works operates programmatically by following if-then rules or other heuristics in order to retrieve, fill and/or combine templates from a database in order to generate a report. The visual component typically outputs discrete classification labels that the language component processes programmatically. In the case of Harzig et al. 2019b \cite{10.1145/3343031.3356066}, image localizations per class are also recovered using CAM \cite{zhou2016learning}, and in the case of Han et al. \cite{10.1007/978-3-030-00937-3_22} the visual component outputs an image segmentation.
In both cases the language component includes special localization-based rules or templates, thus incorporating location information in the generated report.
Kisilev et al. \cite{kisilev2016medical} followed a different approach: a multi-layer perceptron learns to map image encodings to doc2vec \cite{le2014distributed} representations of corresponding reports. During inference, the ground-truth report with the closest doc2vec representation is retrieved.

Lastly, we identify three papers \cite{li2019knowledge, biswal2020clara, li2018hybrid} following the \textit{Hybrid template retrieval + generation/edition} approach. These works seek to combine the benefits of templates with the flexibility of a generative module to either generate sentences from scratch or paraphrase templates as needed on a case-by-case basis. KERP \cite{li2019knowledge} uses \textit{Graph Transformers (GTR)} to map the visual input into a sequence of templates from a curated database. A \textit{Paraphrase} GTR then maps each template to its paraphrased version.
HRGR \cite{li2018hybrid} follows the hierarchical LSTM approach with a twist---it replaces the Word LSTM with a gate module that chooses between two options:
retrieving a template or generating a sentence from scratch (via a Word LSTM).
Lastly, CLARA \cite{biswal2020clara} is somewhat different, as it was designed as an interactive tool to assist a human to write reports. A human introduces \textit{anchor words} and the prefix of a sentence, and \textit{Lucene} \cite{milosavljevic2010retrieval} processes them as a query to retrieve sentence templates from a database. A sequence-to-sequence network then reads and paraphrases each sentence template to get the final report. CLARA can also operate fully automatically by receiving an empty prefix and predicting the anchor words itself. According to reported results, the model consistently achieved better performance than many baselines.

\subsubsection{Domain knowledge}
\label{section:domain-knowledge}

Although all works used datasets from the medical domain to train their models, which can be considered a form of domain knowledge transfer, some works took special steps to explicitly incorporate additional knowledge from experts into their design.
Concretely, we identify two incipient trends in the application of domain knowledge: 1) the use of graph neural networks right after the CNN, providing an architectural bias to guide the model to identify medical concepts and their relations from the images; and 2) enhancing the model's report generation with access to an external template database curated by experts.


KERP \cite{li2019knowledge} incorporates knowledge at the architectural level using graph neural networks. The authors manually designed an abnormality graph and a disease graph, where each node represents an abnormality or disease, and the edges are built based on their co-occurrences in the training set. Some example abnormalities are ``low lung volumes'' and ``enlarged heart size'',
whereas diseases represent a higher level of abstraction, for example ``emphysema'' or ``consolidation''. The information flows from image features (encoded by a CNN) to the abnormality graph, and then to the disease graph,
via inter-node message passing. This biases the network to encode the visual information in terms of abnormalities, diseases and their relations.
Similarly, Zhang et al. \cite{zhang2020radiology} created an observations graph, containing 20 nodes of chest abnormalities or body parts, where conditions related to the same organ or tissue are connected by edges.
Their ablation analysis showed some performance gains, thanks to the graph neural network.

In seven works \cite{li2018hybrid, li2019knowledge, biswal2020clara, 10.1145/3343031.3356066, 10.1007/978-3-030-00937-3_22, kisilev2016medical, moradi2016cross} the authors provided their models with a curated set of template sentences that are further processed in the language component to output a full report.
Three works \cite{10.1007/978-3-030-00937-3_22, 10.1145/3343031.3356066, kisilev2016medical} used manually curated templates and if-then based programs to select and fill them.
CLARA \cite{biswal2020clara} uses a database indexing all sentences from the training set reports for text-based retrieval, which are then paraphrased by a generative module.
Similarly, KERP \cite{li2019knowledge} has access to a template database mined from the training set, which are also paraphrased later.
In HRGR \cite{li2018hybrid} the most common sentences in the datasets were mined and then manually grouped by meaning to further reduce repetitions. In this work the authors showed that HRGR learned to prefer templates about 80\% of the time and only generate sentences from scratch the remaining 20\%, suggesting that templates can be quite useful to generate most sentences in reports.


\subsubsection{Auxiliary Tasks}
\label{section:aux-tasks}

Although the main objective in most papers is to learn a model for report generation from medical images, many works also include and optimize auxiliary tasks to boost their performance. A summary of these tasks is presented in Table \ref{tab:aux-tasks} in appendix \ref{section:appendix:auxtasks}.
%
%
The most common auxiliary tasks are multi-label (16 papers) and single-label (11 papers) classification. These tasks are generally intended to provide additional supervision to the model's visual component, in order to improve the CNN's capabilities to extract quality visual features. Some common tasks are identifying the presence or absence of different abnormalities, diseases, organs, body parts, medical concepts, detecting image modality, etc.
Datasets often used for this purpose are ChestX-ray14 \cite{Wang_2017_CVPR} and CheXpert \cite{irvin2019chexpert}, where the common practice is to pretrain the CNN in those datasets before moving on to report generation. Many papers report better performance in report generation thanks to these auxiliary classification tasks.
The three works \cite{harzig2019addressing, 10.1007/978-3-030-18590-9_64, jing2019show} following the hierarchical approach with Dual Word LSTM used a classification task to supervise the gating mechanism that chooses between generating a normal sentence, an abnormal sentence or stopping.
Two models \cite{10.1007/978-3-030-00934-2_78, 10.1007/978-3-030-00937-3_22} perform a segmentation task. Tian et al. \cite{10.1007/978-3-030-00934-2_78} trained a fully convolutional network (FCN) with segmentation masks of a liver and tumor, and Han et al. \cite{10.1007/978-3-030-00937-3_22} trained an RGAN for pixel level classification.
Similarly, two models  \cite{kisilev2016medical, zeng2020deep} use a Faster-RCNN \cite{ren2015faster} trained for detection and classification of bounding boxes enclosing lesions or other regions of interest in the images.

Two works \cite{maksoud2019coral8, 8970668} used regularization supervision on attention weights. CORAL8 \cite{maksoud2019coral8} receives regularization supervision on its visual attention weights to prevent them from degrading into uniform distribution, which would offer no advantage over average pooling. 
Similarly, Yin et al. \cite{8970668} added two regularizations to their model's attention weights: one on the weights over spatial visual features and another on the weights over tag embedding vectors.
In both works the attention supervision provided a significant contribution to the performance.

Two works \cite{8970668, moradi2016cross} included a task to enforce a matching between embeddings from two different sources.
Yin et al. \cite{8970668} projected the topic vectors from the Sentence LSTM and the word embeddings from the respective ground-truth sentence into a common semantic space, and enforced a matching via contrastive loss \cite{chopra2005learning}. This task significantly improved the Sentence LSTM's training and the model's overall performance.
Moradi et al. \cite{moradi2016cross} trained a MLP for mapping image visual encodings (obtained by a VGG network) to the vector representation of its corresponding ground-truth report (obtained via doc2vec \cite{le2014distributed}, which in itself was another auxiliary task), by minimizing the Euclidean distance. The trained MLP was then used to predict doc2vec representations for unseen images and retrieve the report with the closest representation.
Two works \cite{10.1007/978-3-030-33850-3_8, spinks2019justifying} used text autoencoders, which allow learning compact representations of unlabeled data in a self-supervised manner: an encoder network maps the input into a latent representation, and a decoder network has to recover the original input back.
MTMA \cite{10.1007/978-3-030-33850-3_8} uses a BiLSTM to encode the sentences of the \textit{indication} and \textit{findings} sections of a report (input text), in order to generate the \textit{impression} section (output). To improve the encoding quality of the BiLSTM, the authors trained the decoder branch of a hierarchical autoencoder \cite{li2015hierarchical} to recover the original sentence from the BiLSTM encoding. The experimental results showed that the autoencoder supervision provided a significant boost to the model's performance.
Spinks and Moens \cite{spinks2019justifying} trained an ARAE \cite{zhao2017adversarially} (1) to learn compact representations of reports (serving as input to a GAN that generates chest X-ray images) and (2) to recover a report given an arbitrary compact representation (used in inference mode for report generation).

Lastly, Spinks and Moens \cite{spinks2019justifying} were the only ones to also implement cycle-consistency tasks \cite{zhu2017unpaired} to train a GAN and an inverse mapping CNN together, to make both chest X-ray image generation and encoding more robust.
These tasks will be further detailed in the next section.

\subsubsection{Optimization Strategies}
\label{section:opt-strat}


In addition to the architecture and the tasks a model can perform, a very important aspect is the optimization strategy used to learn the model's parameters. In this section we present an analysis of the optimization strategies used in the literature.  A summary of this section is presented in Table \ref{table:opti} in appendix \ref{section:appendix:optimization}.

\textit{Visual Component}.
We first analyze the visual component optimization, identifying three general optimization decisions. The first one is whether to use a CNN from the literature with its weights pretrained in ImageNet \cite{deng2009imagenet}. This is a very common transfer learning practice from the computer vision literature in general \cite{kornblith2019better}, so it is natural to see it used in the medical domain too.
However, it has been shown that ImageNet pretraining may not transfer as well to medical image tasks as they normally do to other domains, due to very dissimilar image distributions \cite{raghu2019transfusion}. Therefore, a very common second decision
is whether or not to train/fine-tune the visual component with auxiliary medical image tasks, such as most of the classification and segmentation tasks discussed in the previous section (\ref{section:aux-tasks}).
The third decision is whether to freeze the visual component weights during report generation training or continue updating them in an end-to-end manner.


\textit{Report Generation}.
We identify two general optimization strategies in the literature: \textit{Teacher-forcing} (TF) and \textit{Reinforcement Learning} (RL).
Teacher-forcing \cite{williams1989learning} is by far the most common, as it is adopted by 32 papers \cite{jing2017automatic, huang2019multi, yuan2019automatic, li2019knowledge, wang2018tienet, xue2018multimodal, zhang2017mdnet, li2019vispi, singh2019chest, maksoud2019coral8, gale2018producing, 10.1007/978-3-030-00934-2_78, 8935910, 8970668, 10.1007/978-3-030-33850-3_8, 10.1007/978-3-030-32251-9_37, gasimova2019automated, 10.1007/978-981-15-4015-8_15, harzig2019addressing, biswal2020clara, 10.1007/978-3-030-18590-9_64, Zeng2018, 10.1007/978-3-030-20351-1_10, 10.1007/978-3-030-26763-6_66, zhang2020radiology, 10.1145/3357254.3357256, jing2019show, Shin_2016_CVPR, 10.1007/978-3-319-98932-7_21, wu2017generative, spinks2019justifying, zeng2020deep}.
The basic idea in teacher-forcing is to train a model to predict each word of the report conditioned on the previous words, therefore learning to imitate the ground truth word by word.
The model typically has a softmax layer that predicts the next word, and cross entropy is the loss function of choice to measure the error and compute gradients for backpropagation. We think teacher-forcing is so widespread in the literature because of its simplicity and general applicability, as it is agnostic to the application domain (whether it be report generation in medicine or captioning of everyday images).

In contrast, 5 works \cite{liu2019clinically, li2018hybrid, xiong2019reinforced, jing2019show, 10.1145/3357254.3357256} explored the use of reinforcement learning (RL) \cite{kaelbling1996reinforcement}.
The main reason to use RL is the flexibility it offers to optimize non-differentiable reward functions, allowing researchers to be more creative and explore new rewards that may guide the model's learning toward domain-specific goals of interest.
For example, Liu et al. \cite{liu2019clinically} used RL to train their model to optimize the weighted sum of two rewards: (1) a natural language reward (CIDEr \cite{vedantam2015cider}) and (2) a \textit{Clinically Coherent Reward} (CCR), where the latter was proposed to measure the clinical accuracy of a generated report compared to a ground-truth reference using the CheXpert labeler tool \cite{irvin2019chexpert}.
Their goal was to equip their model with two skills: natural language fluency (encouraged by CIDEr) and clinical accuracy (encouraged by CCR).
Other examples of the use of RL are: the direct optimization of CIDEr \cite{li2018hybrid, xiong2019reinforced}, particularly in the training of a complicated hybrid template-retrieval and text generation model \cite{li2018hybrid}; directly optimizing BLEU-4 after a previous teacher-forcing warmup phase \cite{jing2019show}; and the training of the generator network of a GAN used for report generation, where the reward is provided by the discriminator network \cite{10.1145/3357254.3357256}.




As a side note, we would like to highlight the work by Zhang et al. \cite{zhang2019optimizing} on medical report summarization (a related task where the report is the input and with no images), illustrating how RL can be used in this setting to optimize both fluency and factual correctness. As rewards they used ROUGE \cite{lin-2004-rouge} and a Factual Correctness reward based on the CheXpert labeler tool \cite{irvin2019chexpert} (very similar to the CCR proposed by Liu et al. \cite{liu2019clinically}). This work is a good example of the benefits of RL over teacher-forcing for text generation in a medical domain. 
The paper presents the results of a human evaluation with two board-certified radiologists and the model trained with RL achieved better results than the same model trained with teacher-forcing, and even slightly better results than the human baseline.

\textit{Other Losses or Training Strategies}.
This category encompasses the remaining optimization strategies found in the literature.
The most important one is \textit{multitask learning} \cite{caruana1997multitask}, adopted by 14 papers \cite{jing2017automatic, li2019knowledge, wang2018tienet, zhang2017mdnet, maksoud2019coral8, 10.1007/978-3-030-00934-2_78, 8970668, 10.1007/978-3-030-33850-3_8, 10.1007/978-3-030-04224-0_24, harzig2019addressing, jing2019show, kisilev2016medical, spinks2019justifying, zeng2020deep}.
The main idea is to jointly train a model in multiple complementary tasks, so that the model can learn robust parameters that perform well in all of them.
Some works \cite{jing2017automatic, wang2018tienet, zhang2017mdnet, 10.1007/978-3-030-00934-2_78, 8970668, 10.1007/978-3-030-33850-3_8, harzig2019addressing} trained the visual and language components simultaneously in multiple tasks in an end-to-end manner, i.e. report generation plus other auxiliary tasks.
Other examples are the simultaneous training of object detection and attribute classification \cite{kisilev2016medical}, diagnostic classification and cycle-consistency tasks \cite{spinks2019justifying}, among others.
Most of these papers report benefits from training in this way.

As already discussed in section \ref{section:aux-tasks}, two works \cite{maksoud2019coral8, 8970668} used auxiliary supervision on the attention weights of their models. These auxiliary losses were jointly optimized with the rest of the model in report generation, effectively having a regularizer effect.
Yin et al. \cite{8970668} are also the only ones that included an auxiliary contrastive loss \cite{chopra2005learning} to provide a direct supervision to the Sentence LSTM, thus improving their model's performance. Notice that all these works are examples of multitask learning too.
%
%
Three papers \cite{kisilev2016medical, moradi2016cross, zeng2020deep} used regression losses. Two of them \cite{kisilev2016medical, zeng2020deep} included a bounding box regression loss as part of Faster R-CNN \cite{ren2015faster} training, and Moradi et al. \cite{moradi2016cross} included a regression loss to minimize the Euclidean distance between VGG and doc2vec embeddings.
%
As previously discussed in section \ref{section:aux-tasks}, another optimization strategy is the use of autoencoders for the self-supervised learning of text representations. In MTMA \cite{10.1007/978-3-030-33850-3_8} an autoencoder was used to provide an auxiliary supervision over the BiLSTM and was jointly trained with the rest of the model in a multitask learning fashion. Spinks and Moens \cite{spinks2019justifying} instead trained an ARAE in a first stage, then froze its weights and used the learned text embedding to support the subsequent training of a GAN.

Lastly, three works used GANs \cite{10.1007/978-3-030-00937-3_22, 10.1145/3357254.3357256, spinks2019justifying}. As mentioned when discussing RL, Li et al. \cite{10.1007/978-3-030-00937-3_22} used a GAN strategy to train their model for report generation, where the generative module generates a report and the discriminator determines whether it is real or fake. Similarly, Han et al. \cite{10.1145/3357254.3357256} proposed RGAN, where the generator outputs segmentation maps from spine radiographs and the discriminator determines if a given segmentation map is real or fake. Spinks and Moens \cite{spinks2019justifying} implemented a modified version of a StackGAN \cite{zhang2017stackgan} to generate chest X-ray images from input text representations. In their case, they trained the GAN using two cycle-consistency \cite{zhu2017unpaired} losses: (1) image $\xrightarrow{}$ embedding $\xrightarrow{}$ image and (2) embedding $\xrightarrow{}$ image $\xrightarrow{}$ embedding. In both cases, an auxiliary inverse mapping CNN was used to close the cycle.

\vspace{-2mm}
\subsubsection*{Synthesis}
Overall, we can observe that designing a model for report generation from medical images is a complex task that involves engineering decisions at multiple levels: inputs and outputs, visual component, language component, domain knowledge, auxiliary tasks and optimization strategies. In each of these dimensions there are different approaches adopted in the reviewed literature, and the current state of research does not allow us to recommend an ``optimal model design'', mainly for reasons we will discuss in the Metrics and Performance Comparison sections (\ref{section:metrics} and \ref{section:comparison-papers}).
Nevertheless, there are valuable insights in the literature that may lead to better results, and thus are worth having in mind. For example, the use of CNNs (such as DenseNet or ResNet) as visual component and training in auxiliary medical image tasks;
the use of input text alongside the images;
providing the language component with tag information in addition to the visual features (e.g. medical concepts identified in the image);
leveraging template databases curated with domain knowledge; or
the use of multitask learning combining multiple sources of supervision.
Lastly, to improve report quality from a medical perspective, the use of reinforcement learning with adequate reward functions appears as the most promising approach.


\vspace{-3mm}
\subsection{Explainability}
\label{section:xai}

There have been multiple attempts on providing a definition for \textit{explainability} in the Explainable Artificial Intelligence (XAI) area \cite{reyes2020interpretability, doshi2017towards, lipton2018mythos}.
For the task of report generation from medical images, we use a similar definition by Doshi-Velez and Kim \cite{doshi2017towards}: \textit{the ability to justify an outcome in understandable terms for a human}, and we use it interchangeably with the term \textit{interpretability}. 
In this medical context, an automated system requires high explainability levels as two main facts hold: the decisions derived from the system will probably have direct consequences for patients, and the diagnosis task is not trivial and susceptible to human judgement \cite{reyes2020interpretability, doshi2017towards}.
Furthermore, the explanation methods employed in this medical task should attempt to solve several related aspects:
align with clinicians' expectations and acquire their trust, 
increase system transparency, 
assess results quality, 
and allow addressing accountability, fairness and ethical concerns
\cite{ahmad2018interpretable, reyes2020interpretability, tonekaboni2019clinicians}.

There are many ways to address the explainability aspect of AI systems in the medical domain, as listed in the recent survey on interpretable AI for radiology by Reyes et al. \cite{reyes2020interpretability}.
Multiple categories can be identified, starting with \textit{global vs local}, the former refers to explanations regarding the whole system's operation, and the latter to explanations for one sample.
For local explanations, there are different kinds of approaches, such as \textit{feature importance}, \textit{concept-based}, \textit{example-based}, and \textit{uncertainty}, to mention a few.
\textit{Feature importance} methods attempt to compute a level of importance for each input value, to understand which characteristics were most relevant to make a decision; for example, gradient-based methods for CNNs such as Grad-CAM \cite{selvaraju2016grad}, Guided Backpropagation \cite{springenberg2015striving} or DeepLIFT \cite{shrikumar2017deeplift}; and other techniques such as LIME \cite{ribeiro2016lime}.
In \textit{concept-based} methods, like TCAV \cite{kim2017interpretability} or RCV \cite{graziani2018rcv}, the contributions to the prediction from multiple concepts are quantified, so the user can check if the concepts used by the model are correct.
\textit{Example-based} approaches present additional examples with the output, either with a similar outcome, so the user can look for a common pattern, or with an opposite outcome (counter-factual).
\textit{Uncertainty} methods provide the level of confidence of the model for a given prediction.
For global explanations, there are \textit{sample-based} approaches, such as SP-LIME \cite{ribeiro2016lime}, or methods to directly increase the \textit{transparency} of the system.


Despite the importance of explainability in this area,
only two reviewed works focused explicitly on this topic.
Gale et al. \cite{gale2018producing} proposed the automatic generation of a natural language report as an explanation for a classification task; however, their approach does not include an explanation for the report.
Spinks and Moens \cite{spinks2019justifying} present a counter-factual local explanation, as will be detailed in subsection \ref{section:xai:counter-factual}.
Additionally, in 29 works the model architecture generates a secondary output that can also be presented as a local explanation.
We distinguish three types of outputs:
classification (section \ref{section:xai:classification}),
heatmap over the input image (section \ref{section:xai:image-heatmap}),
and heatmap over the input text (section \ref{section:xai:text-heatmap}).
These were already summarized in Table \ref{table:io} in the Input and Output section (\ref{section:input-output}).
Next, the explanaibility aspects of the outputs are discussed.

\subsubsection{Counter factual}
\label{section:xai:counter-factual}

Spinks and Moens \cite{spinks2019justifying} proposed an architecture to both classify a disease and generate a caption from a chest X-ray, based on GANs and autoencoders, as detailed in the Model Design section (\ref{section:model-design}).
Thus, to provide a local explanation, at inference time the input image is encoded into a latent vector, which is used to generate a new chest X-ray and a new report, both of them subject to result in the nearest alternative classification, i.e., the nearest diagnosis.
With this information, a user could compare the original X-ray with the generated image, and attempt to understand why the model has reached its decision.

\subsubsection{Classification}
\label{section:xai:classification}

As explained in the Auxiliary Tasks section (\ref{section:aux-tasks}), many deep learning architectures include multi-label classification to improve performance, providing a set of classified concepts as secondary output.
Even though in most papers this kind of output is not presented as an explanation of the report, we consider that its nature could improve the transparency of the model, which is an important way of improving the interpretability in a medical context \cite{tonekaboni2019clinicians}.
By providing this detection information from an intermediate step of the model's process, an expert could further understand the internal process, validate the decision with their domain knowledge and calibrate their trust in the system.

As shown in Table \ref{table:io} from section \ref{section:input-output}, the terms classified are very diverse.
Some works classify very broad concepts, such as body parts or organs \cite{Zeng2018, 10.1007/978-3-030-32251-9_37, moradi2016cross, zeng2020deep}, or image modality \cite{10.1007/978-3-319-98932-7_21}.
Other works perform a more specific classification, such as diseases or abnormalities \cite{Zeng2018, wang2018tienet, biswal2020clara, li2019knowledge, zhang2020radiology, zeng2020deep, spinks2019justifying, kisilev2016medical}, or a normal or abnormal status at sentence level \cite{10.1007/978-3-030-18590-9_64}.
Lastly, several works \cite{harzig2019addressing, jing2017automatic, 10.1007/978-3-030-33850-3_8, 8935910, 8970668, 10.1007/978-3-030-26763-6_66, yuan2019automatic, Shin_2016_CVPR} classify over a subset of MeSH terms or similar, which may contain a mix of general broad medical concepts and specific abnormalities or conditions.
We believe that this additional output would be useful for an expert, though the specific concepts should provide much richer information.
If the classification is more specific, the user will be able to validate on a much narrower scope the system's performance.

\subsubsection{Image heatmap}
\label{section:xai:image-heatmap}

In the papers reviewed, there are three different approaches to generating heatmaps over the input image, each of them with a different interpretation.
First, many architectures employ an attention mechanism over the image spatial features during the report generation, as it was discussed in the Language Component section (\ref{section:language-component}).
These mechanisms can be leveraged to produce a heatmap indicating the image regions that were most important to generate the report.
In particular,
some models provide a heatmap for each word \cite{zhang2017mdnet, wang2018tienet, liu2019clinically}, 
for each sentence \cite{jing2017automatic, huang2019multi, xue2018multimodal, 10.1007/978-3-030-20351-1_10},
or for the whole report \cite{li2019knowledge}.
By showing these feature importance maps, an expert should be able to determine if the model is focusing on the correct regions of the image, which could improve their trust on the system.



Second, some works
use particular deep learning architectures to perform image segmentation, i.e. classification and localization at the same time.
The model by Ma et al. \cite{10.1007/978-3-030-04224-0_24} uses a CNN to classify the severity of four different key characteristics of cervical cancer, and then uses an attention mechanism over the visual spatial features to generate heatmaps indicating the position of each relevant property.
Tian et al. \cite{10.1007/978-3-030-00934-2_78} used an FCN to classify each pixel of an image with the presence of a liver or tumor, and the result is averaged with an attention map to further improve localization.
Han et al. \cite{10.1007/978-3-030-00937-3_22} proposed the ACAE module (see section \ref{section:language-component} for details), which is used to classify at pixel level different parts of the spine (vertebrae, discs or neural foramina), and if they show an abnormality or not.
Kisilev et al. \cite{kisilev2016medical} and Spinks and Moens \cite{spinks2019justifying} used a Faster R-CNN \cite{ren2015faster} architecture to detect image regions with lesion and body parts of interest.

\begin{figure}[tpb]
    \centering
    \begin{tabular}{cc}
    \includegraphics[width=0.23\linewidth]{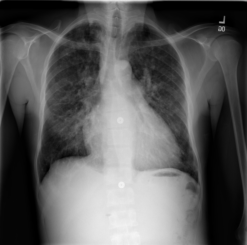}
    \includegraphics[width=0.23\linewidth]{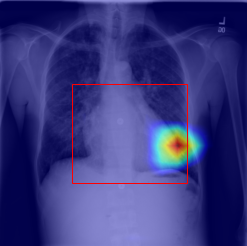}
    &
    \includegraphics[width=0.23\linewidth]{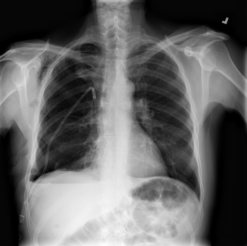}
    \includegraphics[width=0.23\linewidth]{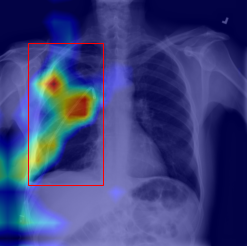}
    \\
    \small{Cardiomegaly}
    &
    \small{Pneumothorax}
    \end{tabular}
    \vspace{-3mm}
    \caption{Examples from the ChestX-ray14 dataset \cite{Wang_2017_CVPR} classified with a CNN based on ResNet-50 \cite{he2016deep}, and using CAM \cite{zhou2016learning} to provide a heatmap indicating the the spatial regions of most importance as local explanation.
    The left example presents Cardiomegaly and the right Pneumothorax, and both samples were correctly classified by the CNN.
    Red boxes represent a localization of the condition annotated by an expert.
    }
    \Description{Two chest X-rays showing abnormalities, alongside a heatmap indicating the regions of most importance for the neural network, plus a bounding-box locating the abnormality.}
    \label{figure:cam-example}
    \vspace{-6mm}    
\end{figure}

Lastly, some works use gradient- or activation-based methods for CNNs to generate a saliency map indicating the regions of most importance for a classification, such as
CAM \cite{zhou2016learning},
Grad-CAM \cite{selvaraju2016grad},
SmoothGrad \cite{smilkov2017smoothgrad},
or the one proposed by Zagoruyko and Komodakis \cite{zagoruyko2016paying}.
Refer to Table \ref{table:io} in the Input and Output section (\ref{section:input-output}) for a list of the papers using each technique.
To determine which of these methods performs better in a general setting, Adebayo et al. \cite{adebayo2018sanity} performed multiple evaluations (``sanity checks'') over Grad-CAM, SmoothGrad, and other similar methods, and showed that Grad-CAM should be more reliable in terms of correlation with the input images and the classification made.
As an example of these techniques, Figure \ref{figure:cam-example} shows two chest X-rays from the ChestX-ray14 dataset \cite{Wang_2017_CVPR} with a heatmap generated with CAM, plus an expert-annotated bounding box locating the abnormality (provided with the dataset).

In both segmentation and saliency map methods, the heatmap information provides much richer information than classification alone, as it also includes the location of an specific concept, such as an abnormality or a body part.
Providing this type of explanation should allow an expert to assess the localization capabilities of the model and the system accuracy, thus improving the model's transparency throughout its process. 

\subsubsection{Text heatmap}
\label{section:xai:text-heatmap}

The model proposed by Huang et al. \cite{huang2019multi} also receives text as input, which indicates the reason for performing the imaging study on the patient.
In a similar fashion to the input image cases, the architecture includes an attention mechanism over the input text, which provides a heatmap indicating the input phrases or sentences that were most relevant to generate each word in the output.
With this feature importance map an expert should be able to determine if the model is focusing on the correct words in the input text.

\subsubsection*{Synthesis}

All the explainability approaches are local explanations given by a secondary output, either indicating feature importance (image and text heatmap), increasing the model's transparency (classification) or providing a counter-factual example.
However, in most of the works the authors do not explicitly mention it as an interpretability improvement, and in almost all cases there is no formal evaluation, as will be discussed in subsection \ref{section:metrics:xai}.
Hence, we believe this is an understudied aspect of the medical report generation task, given the superficial or nonexistent analysis it receives in most of the reviewed works.
Additionally, counter-factual techniques could be further studied, and other approaches not found in the literature could be explored, such as prediction uncertainty or global explanations, which may be quite relevant for clinicians \cite{tonekaboni2019clinicians}.

\subsection{Evaluation Metrics}
\label{section:metrics}

There are different ways to assess a medical report generated by an automated system.
We divide the evaluation metrics used in the literature into three categories, depending on the aspect being assessed: text quality, medical correctness and explainability.
Also, each evaluation method can be either automatic or performed manually by humans.
Each of the categories and metrics are presented next, and Table \ref{table:metrics-summary} shows a summary of the metrics used by each paper.

\begin{table}[htbp]
\scalebox{0.78}{
\begin{tabular}{|p{0.28\textwidth}|p{0.65\textwidth}|p{0.27\textwidth}|}
\hline
 \textbf{Category} & \textbf{Metric or evaluation} & \textbf{Used by papers}
 \\ \hline
 \multirow[t]{7}{\linewidth}{Text quality (automatic)}
 & BLEU based & \raggedright{\cite{jing2017automatic, liu2019clinically, huang2019multi, yuan2019automatic, li2018hybrid, li2019knowledge, wang2018tienet, xue2018multimodal, zhang2017mdnet, li2019vispi, xiong2019reinforced, singh2019chest, maksoud2019coral8, gale2018producing, 10.1007/978-3-030-00934-2_78, 8935910, 8970668, 10.1007/978-3-030-33850-3_8, 10.1007/978-3-030-32251-9_37, gasimova2019automated, 10.1007/978-981-15-4015-8_15, harzig2019addressing, biswal2020clara, 10.1007/978-3-030-18590-9_64, Zeng2018, 10.1007/978-3-030-20351-1_10, 10.1007/978-3-030-26763-6_66, zhang2020radiology, 10.1145/3357254.3357256, jing2019show, Shin_2016_CVPR, 10.1007/978-3-319-98932-7_21, spinks2019justifying, zeng2020deep}} \tabularnewline 
 & ROUGE-L & \raggedright{\cite{jing2017automatic, liu2019clinically, huang2019multi, yuan2019automatic, li2018hybrid, li2019knowledge, wang2018tienet, xue2018multimodal, zhang2017mdnet, li2019vispi, singh2019chest, maksoud2019coral8, 10.1007/978-3-030-00934-2_78, 8935910, 8970668, 10.1007/978-3-030-33850-3_8, 10.1007/978-3-030-32251-9_37, 10.1007/978-981-15-4015-8_15, harzig2019addressing, 10.1007/978-3-030-18590-9_64, Zeng2018, 10.1007/978-3-030-20351-1_10, zhang2020radiology, jing2019show, spinks2019justifying, zeng2020deep}} \tabularnewline
 & METEOR based & \raggedright{\cite{jing2017automatic, yuan2019automatic, wang2018tienet, xue2018multimodal, zhang2017mdnet, singh2019chest, maksoud2019coral8, 8935910, 8970668, 10.1007/978-981-15-4015-8_15, harzig2019addressing, Zeng2018, 10.1007/978-3-030-20351-1_10, 10.1145/3357254.3357256, spinks2019justifying, zeng2020deep}} \tabularnewline
 & CIDEr based & \raggedright{\cite{jing2017automatic, liu2019clinically, huang2019multi, li2018hybrid, li2019knowledge, zhang2017mdnet, li2019vispi, xiong2019reinforced, singh2019chest, 8970668, 10.1007/978-981-15-4015-8_15, harzig2019addressing, biswal2020clara, 10.1007/978-3-030-18590-9_64, Zeng2018, 10.1007/978-3-030-20351-1_10, 10.1007/978-3-030-26763-6_66, zhang2020radiology, jing2019show, spinks2019justifying, zeng2020deep}} \tabularnewline
 & SPICE & \cite{10.1145/3357254.3357256} \\
 & Grammar Bot & \cite{10.1007/978-3-030-32251-9_37} \\
 & Sentence variability & \cite{harzig2019addressing} \\
 \hline
 Text quality (with humans)
 & AMT study & \cite{li2018hybrid, li2019knowledge}\\
 \hline
 \multirow[t]{8}{\linewidth}{\raggedright{Medical correctness (automatic, report based)}}
 & MIRQI (precision, recall, F1) & \cite{zhang2020radiology} \\ 
 & MeSH Accuracy & \cite{huang2019multi} \\
 & Keyword ratio (accuracy, sensitivity, specificity) & \cite{wu2017generative} \\
 & Keyword Accuracy & \cite{xue2018multimodal, 10.1007/978-3-030-18590-9_64} \\
 & \raggedright{ Medical Abnormality Terminology Detection (precision, FPR)} & \cite{li2018hybrid} \\
 & Abnormality Detection (precision, FPR) & \cite{jing2019show} \\
 & Medical Abnormality Detection (accuracy, precision, recall) & \cite{liu2019clinically} \\ 
 & Abnormality CNN classifier (accuracy, PR-AUC)  & \cite{biswal2020clara} \\
 & Semantic descriptors & \cite{moradi2016cross} \\
 & ARS & \cite{10.1007/978-3-030-32251-9_37} \\
 \hline
 \multirow[t]{10}{\linewidth}{\raggedright{Medical correctness (automatic, auxiliary tasks)}}
 & ROC-AUC & \cite{li2019knowledge, wang2018tienet, li2019vispi, zhang2020radiology}\\
 & Accuracy & \cite{zhang2017mdnet, Zeng2018, Shin_2016_CVPR, 10.1007/978-3-030-04224-0_24, kisilev2016medical, zeng2020deep} \\
 & Recall/sensitivity & \cite{8970668, 10.1145/3343031.3356066} \\
 & Precision & \cite{8970668, 10.1145/3343031.3356066, kisilev2016medical} \\
 & Specificity & \cite{10.1145/3343031.3356066} \\
 & Pixel level accuracy & \cite{10.1007/978-3-030-00937-3_22} \\
 & Pixel level specificity & \cite{10.1007/978-3-030-00937-3_22} \\
 & Pixel level sensitivity & \cite{10.1007/978-3-030-00937-3_22} \\ 
 & Pixel level dice score & \cite{10.1007/978-3-030-00937-3_22, 10.1007/978-3-030-00934-2_78} \\
 \hline
 \multirow[t]{2}{\linewidth}{\raggedright{Medical correctness (with experts)}}
 & Assess correctness of the nature of hip fractures & \cite{gale2018producing}\\
 & Accept/reject rating & \cite{10.1007/978-3-030-00934-2_78} \\
 & Assess medical and grammatical correctness, and relevance & \cite{10.1007/978-3-030-32251-9_37} \\
 & Agree with diagnosis & \cite{spinks2019justifying} \\
 \hline
 \multirow[t]{2}{\linewidth}{\raggedright{Explainability (with experts)}}
 & Counter factual X-ray vs Saliency map & \cite{spinks2019justifying} \\
 & Reports vs SmoothGrad (classification explanation) & \cite{gale2018producing} \\
 \hline
\end{tabular}
}
\caption{Summary of the evaluation metrics used in the literature. The report based medical correctness type includes metrics that are measured from the report generated; the auxiliary task medical correctness ones evaluate an auxiliary or intermediate task in the process, such as classification or segmentation.}
\label{table:metrics-summary}
\vspace{-10mm}
\end{table}

\subsubsection{Text quality metrics}
\label{section:metrics:text-quality}

The methods in this category measure general quality aspects of the generated text, and are originated from translation, summarizing or captioning tasks.
The most widely used metrics in the papers reviewed are
BLEU \cite{papineni2002bleu},
ROUGE-L \cite{lin-2004-rouge},
METEOR \cite{banerjee2005meteor, 10.5555/1626355.1626389}
and CIDEr \cite{vedantam2015cider},
which measure the similarity of a target text (also referred to as candidate), against one or more reference texts (ground truth).
These metrics are mainly based on counting n-gram matchings between the candidate and the ground truth.
BLEU is precision-oriented, ROUGE-L and METEOR are F1-scores that can be biased towards precision or recall with a given parameter, and CIDEr attempts to capture both precision and recall through a TF-IDF score.
Most of these metrics have variants and parameters for their calculation:
ROUGE is a set of multiple metrics, being ROUGE-L the only one used in this task;
METEOR has variants presented by the same authors \cite{10.5555/1857999.1858030, 10.5555/2132960.2132969, denkowski2014meteor};
and CIDEr was presented with the CIDEr-D variant to prevent gameability effects.

SPICE \cite{10.1007/978-3-319-46454-1_24} is a metric designed for the image captioning task, and evaluates the underlying meaning of the sentences describing the image scene, partially disregarding fluency or grammatical aspects.
Specifically, the text is parsed as a graph, capturing the objects, their described characteristics and relations, which are then measured against the ground truth using an F1-score.
Even though SPICE attempts to assess the semantic information in a caption, we believe it is not suitable for medical reports, as the graph parsing is designed for general domain objects.
Nonetheless, Zhang et al. \cite{zhang2020radiology} presented the medical correctness metric MIRQI, applying a similar idea in a specific medical domain, which we will discuss in the next subsection (\ref{section:metrics:med-corr}).

Besides standard captioning metrics, we identified two other approaches to measure text quality.
First, Alsharid et al. \cite{10.1007/978-3-030-32251-9_37} used Grammar Bot\footnote{\url{https://www.grammarbot.io/}}, a rule and statistics based automated system that counts the grammatical errors in sentences.
Second, Harzig et al. 2019a \cite{harzig2019addressing} measured the sentence variability, by counting the different sentences in a set of reports.
They argue that the sentences indicating abnormalities occur very rarely in the dataset, while the ones indicating normality are the most frequent.
Hence, a certain level of variability is desired, and a system generating reports with low variability may indicate that not all medical conditions are being captured.

Lastly, both works from Li et al. \cite{li2018hybrid, li2019knowledge} performed human evaluation with non-expert users via Amazon Mechanical Turk (AMT), following the same procedure.
The authors presented two reports to the AMT participants, one generated with the proposed model and one generated with the CoAtt model \cite{jing2017automatic} as baseline, and asked them to choose the most similar with the ground truth in terms of fluency, abnormalities correctness and content coverage.
The results shown that their report was preferred around 50-60\% of the cases, while the baseline around 20-30\% (for the rest, none or both were preferred).
We categorize this evaluation as a text quality metric, as the participants are not experts, and their answers are not fine-grained (i.e., did not specify what failed: fluency, correctness or coverage; or by how much they failed).

\vspace{-3mm}
\subsubsection{Medical correctness metrics}
\label{section:metrics:med-corr}

While the most common purpose of the text quality metrics is to measure the similarity between the generated report and a ground truth, they do not necessarily capture the medical facts in the reports \cite{boagbaselines,zhang2019optimizing, liu2019clinically,babar2021evaluating,pino2021clinically,pino2020inspecting}.
For example, the sentences \textit{``effusion is observed''} and \textit{``effusion is not observed''} are very similar, thus may present a very high score for any metric based on n-gram matching, though the medical facts are the exact opposite.
Therefore, an evaluation directly measuring the reports correctness is required, not necessarily taking into account fluency, grammatical rules or text quality in general.
From the literature reviewed, in ten works \cite{huang2019multi, xue2018multimodal, li2018hybrid, jing2019show, liu2019clinically, biswal2020clara, 10.1007/978-3-030-32251-9_37, zhang2020radiology, moradi2016cross, wu2017generative} the authors presented an automatic metric to address this issue,
four works \cite{gale2018producing, 10.1007/978-3-030-00934-2_78, 10.1007/978-3-030-32251-9_37, spinks2019justifying} did a formal expert evaluation,
and multiple works \cite{yuan2019automatic, li2019knowledge, wang2018tienet, zhang2017mdnet, li2019vispi, 8970668, 10.1007/978-3-030-04224-0_24, Zeng2018, 10.1145/3343031.3356066, zhang2020radiology, Shin_2016_CVPR, 10.1007/978-3-030-00934-2_78, zeng2020deep, spinks2019justifying, kisilev2016medical} evaluated medical correctness indirectly from auxiliary tasks.
The methods are listed in Table \ref{table:metrics-summary} and are further discussed next.

In several works the authors presented a method that detects concepts in the generated and ground truth reports, and compare the results using common classification metrics, such as accuracy, F1-score, and more.
The main difference between these methods lies in how the concepts are automatically detected in the reports.
The simplest approaches are \textit{keyword-based}, which consists in reporting the ratio of a set of keywords found between the generated report and ground truth,
like MeSH Accuracy \cite{huang2019multi} that uses MeSH terms, and Keyword Accuracy that uses 438 MTI terms (presented by Xue et al. \cite{xue2018multimodal} and used in A3FN \cite{10.1007/978-3-030-18590-9_64}).
Similarly, Medical Abnormality Terminology Detection \cite{li2018hybrid} calculates precision and false positive rate of the 10 most frequent abnormality-related terms in the dataset;
and Wu et al. \cite{wu2017generative} calculated accuracy, sensitivity and specificity for a set of keywords.

Other approaches are \textit{abnormality-based}, which attempt to directly classify abnormalities from the report by different means: Abnormality Detection \cite{jing2019show} uses manually designed patterns;
Medical Abnormality Detection \cite{liu2019clinically} uses the CheXpert labeler tool \cite{irvin2019chexpert};
Biswal et al. \cite{biswal2020clara} used a character-level CNN \cite{zhang2015character} that classifies multiple CheXpert labels \cite{irvin2019chexpert}; and
Moradi et al. \cite{moradi2016cross} used a proprietary software to extract semantic descriptors.
Lastly, Anatomical Relevance Score (ARS) \cite{10.1007/978-3-030-32251-9_37} is a \textit{body-part-based} approach, which detects the anatomical elements mentioned in a report considering the vocabulary used. 
Though these methods may be useful for measuring medical correctness to a certain degree, there is no consensus or standard, and there is no formal evaluation of the correlation with expert judgement.
From the discussed techniques, Alsharid et al. \cite{10.1007/978-3-030-32251-9_37} are the only authors that also performed an expert evaluation of the generated reports, though they did not conduct a correlation or similar analysis to validate the ARS method.

Zhang et al. \cite{zhang2020radiology} went further with the concept extraction and presented Medical Image Report Quality Index (MIRQI), which works in a similar fashion as the SPICE \cite{10.1007/978-3-319-46454-1_24} metric presented in the text quality subsection (\ref{section:metrics:text-quality}).
MIRQI applies ideas from NegBio \cite{peng2018negbio} and the CheXpert labeler \cite{irvin2019chexpert} to identify diseases or medical conditions in the reports, considering synonyms and negations,
and uses the Stanford parser \cite{chen2014fast} to obtain semantic dependencies and finer-grained attributes from each sentence, such as severity, size, shape, body parts, etc.
With this information, an abnormality graph is built for each report, where each node is a disease with its attributes, and the nodes are connected if they belong to the same organ or tissue.
Lastly, the graphs from the ground truth and generated reports are matched node-wise, and MIRQI-p (precision), MIRQI-r (recall) and MIRQI-F1 (F1-score) are computed.
Compared to the formerly discussed correctness metrics, we believe this approach seems more robust to assess the medical facts in the reports, as it attempts to capture the attributes and relations, opposed to the concepts only.
However, the authors did not present an evaluation against expert judgement, so we cannot determine if this metric is sufficient.

Considering human evaluation, only a few works
\cite{gale2018producing, 10.1007/978-3-030-00934-2_78, 10.1007/978-3-030-32251-9_37, spinks2019justifying}
present a formal expert medical correctness assessment.
In the work by Alsharid et al. \cite{10.1007/978-3-030-32251-9_37} a medical professional assessed the reports on a Likert Scale from 0 to 2 in four different aspects: \textit{accurately describes the image}, \textit{presents no incorrect information}, \textit{is grammatically correct} and \textit{is relevant for the image};
the results were further separated for samples from different body parts, showing averages between 0.5 and 1.
Gale et al. \cite{gale2018producing} asked a radiologist to evaluate the correctness of the hip fractures description, finding that the fracture's character was properly described 98\% of the cases, while the fracture location only for 90\%.
In the work by Tian et al. \cite{10.1007/978-3-030-00934-2_78} a medical expert evaluated 30 randomly selected reports with a rating from 1 (\textit{definite accept}) to 5 (\textit{definite reject}), scoring an average of 2.33.
Lastly, Spinks and Moens \cite{spinks2019justifying} asked four questions to three experts regarding the generated reports, where the third and fourth questions measured correctness: \textit{``Do you agree with the proposed diagnosis?''}, answering 0 (\textit{no}) or 1 (\textit{yes}) and \textit{``How certain are you about your final diagnosis?''}, from 1 (\textit{not sure}) to 4 (\textit{very sure}).
The average scores were high (0.88 and 3.75), showing agreement with the model's diagnosis, and certainty on the experts' diagnoses.
The other questions concerned explainability aspects, and are detailed in the next subsection (\ref{section:metrics:xai}).
So far, there is no standard approach to perform an expert evaluation, though we believe the first two approaches provide finer-grained information than the latter two, and hence should be more useful for determining in which cases the models are failing and for designing improvements.
The certainty question should also be very useful, as diagnoses may be susceptible to human judgement.

Lastly, multiple papers \cite{yuan2019automatic, li2019knowledge, wang2018tienet, zhang2017mdnet, li2019vispi, 8970668, 10.1007/978-3-030-04224-0_24, Zeng2018, 10.1145/3343031.3356066, zhang2020radiology, Shin_2016_CVPR, 10.1007/978-3-030-00934-2_78}
evaluated the performance of the auxiliary tasks with ROC-AUC, accuracy and other typical classification or segmentation metrics, as shown in Table \ref{table:metrics-summary}.
Note that in any of these cases, the task is a previous or intermediary step of the process and is not derived from the report.
In consequence, even if the classification has great performance, the language component could be performing poorly, and the generated reports still may be inaccurate.
Accordingly, we believe this type of measure should not be used as the primary report correctness evaluation, unless it can be proven that the report reproduces exactly the classification made (e.g. by a template-filling process).

\vspace{-3mm}
\subsubsection{Explainability metrics}
\label{section:metrics:xai}

Providing interpretable justifications for the model's outcome is essential in this medical domain,
and furthermore, we should be able to evaluate them to answer questions such as,
does the method justify the model's decision?,
which method provides a \textit{better} explanation?
However, there is no consensus on evaluation methods for AI explainability, and in many cases the definition of a \textit{better} explanation remains subjective \cite{doshi2017towards, reyes2020interpretability, carvalho2019machine}.

Consequently, none of the papers reviewed used an automatic metric to assess explainability, and only two works \cite{gale2018producing, spinks2019justifying} conduct a formal human expert evaluation.
Gale et al. \cite{gale2018producing} presented the report generation as an explanation of a medical image classification task, and evaluated it by comparing three methods: (a) SmoothGrad \cite{smilkov2017smoothgrad} to highlight the most important pixels used,
(b) a generated report in natural language,
and (c) both placed side by side.
Five experts assessed 30 images, rating each explanation in a scale from 1 (\textit{unsatisfactory}) to 10 (\textit{perfect}); achieving average scores of (a) 4.4, (b) 7 and (c) 8.8 for each method.
Though the authors emphasize the importance of the natural language explanations, their approach does not include an explanation for the report itself, so it cannot be directly used for the report generation task.

The model proposed by Spinks and Moens \cite{spinks2019justifying} generates a chest X-ray as a counter-factual example,
and they compared this explanation method against a feature importance heatmap generated with the Zagoruyko and Komodakis saliency map technique \cite{zagoruyko2016paying}.
Three experts evaluated 150 samples answering four questions, the first two regarding explainability aspects:
\textit{``Does the explanation justify the diagnosis?''}
\textit{``Does the model appear to understand the important parts of the X-ray?''}
The answers were in a scale from 1 (\textit{no}) to 4 (\textit{yes}), and their method achieved a higher score than the saliency map (2.39 vs 1.31 for the first question, and 2.45 vs 1.81 for the second), showing their counter-factual approach should be \textit{better} in this setting.
The other two questions relate more to medical correctness, and are discussed in the previous section (\ref{section:metrics:med-corr}).

We believe the explanation evaluations should be very important in this area, and as there is no consensus, we outline some possible guidelines.
Following ideas from Tonekaboni et al. \cite{tonekaboni2019clinicians}, we believe three aspects from the explanations should be assessed: (1) consistency, (2) alignment with domain knowledge, and (3) user impact.
First, the consistency across the data should be assessed, answering questions such as:
    do explanations change with variations to the input data?;
    or to the prediction?;
    or to the model design?;
    or with different images from the same patient?
As pointed out by Tonekaboni et al. \cite{tonekaboni2019clinicians}, inconsistent explanations may negatively affect the clinicians' trust, and an interpretability method laying them out should be reviewed.
Examples of consistency or robustness evaluations can be found in the work by Adebayo et al. \cite{adebayo2018sanity} for image saliency maps, and in the work by Jain and Wallace \cite{jain2019attention} for attention in recurrent neural networks.

Second, the alignment with domain knowledge should evaluate if the explanation is consistent with an expert's knowledge:
would they provide the same explanation for that decision?
For instance, given a feature importance method, is the model focusing on the correct features?
As an example, consider the second and third questions employed by Spinks and Moens \cite{spinks2019justifying} detailed earlier.
To mention other examples, Wang et al. \cite{Wang_2017_CVPR} evaluated CAM \cite{zhou2016learning} generated heatmaps for disease classification against expert provided bounding-boxes locating the diseases, using \textit{intersection-over-union} like metrics; Kim et al. \cite{kim2017interpretability} proposed a model to classify Diabetic Retinopathy from retina fundus images, and they compared the TCAV \cite{kim2017interpretability} extracted concepts against expert knowledge.
Notice many works reviewed in this survey used classification or segmentation as an auxiliary task, which can be used as local explanations, and evaluated them with common metrics (such as accuracy, precision, etc.), as discussed in the previous sections (\ref{section:xai} and \ref{section:metrics:med-corr}).
As the authors did not mention the secondary outputs as local explanations, we categorized the said evaluations as \textit{medical correctness} metrics, but they are also measuring \textit{alignment with domain knowledge} for the interpretability methods, and as such may be very useful.

Lastly, the user impact should attempt to answer questions like,
is it a \textit{good} explanation?
Does it provide \textit{useful} or \textit{novel} information?
Does it justify the model's decision?
Is it provided with an appropriate representation for the experts?
As examples, the assessment proposed by Gale et al. \cite{gale2018producing} and the first question used by Spinks and Moens \cite{spinks2019justifying} measure user impact.
Notice that most of these concepts are very subjective, and the definitions, the questions and assessments will vary for different sub-domains and target experts.
We believe more specific definitions and fine-grained aspects should arise in the future, as research in this topic grows.
For reference, this category includes the \textit{domain appropriate representation} and \textit{potential actionability} concepts presented by Tonekaboni et al. \cite{tonekaboni2019clinicians}.

\vspace{-3mm}
\subsubsection*{Synthesis}

Almost all the works include text quality metrics, though these are not able to capture the medical facts in a report \cite{boagbaselines,zhang2019optimizing, liu2019clinically,babar2021evaluating,pino2021clinically,pino2020inspecting}.
Several works proposed medical correctness assessments over the reports, but unfortunately none of the proposals was evaluated against expert judgement.
The auxiliary tasks can be evaluated to measure correctness indirectly from the process, but often it will not be sufficient for the report's correctness.
Only two works 
evaluate explainability directly with experts, and the auxiliary tasks' assessments could be useful to measure alignment between the explanations and domain knowledge. 
Overall, we believe that medical correctness should be the primary aspect to evaluate in the generated reports, using one or more automatic metrics.
For now, and even though none of the metrics proposed has been evaluated against expert judgement, MIRQI \cite{zhang2020radiology} seems like the most promising approach to fulfill this purpose, as it should be able to capture richer information from the reports.
Additionally, text quality metrics can be used as a secondary evaluation, since they may be useful for measuring fluency, grammar or variability, and to compare with previous baselines.
Lastly, explainability evaluation methods should arise to assess multiple key aspects,
such as its consistency, alignment with domain knowledge, and the user impact.

\vspace{-3mm}
\subsection{Comparison of papers' performance}
\label{section:comparison-papers}

To find out which paper holds the state of the art, we need to find a common ground for fair comparison. A natural choice is the IU X-ray dataset \cite{10.1093/jamia/ocv080}, since a majority of the surveyed papers report results in this dataset.
Table \ref{table:iu-xray-results} shows these results, separated by which report sections are generated by each paper, \textit{findings}, \textit{impression} or both.
The findings section consists of multiple sentences, and mainly describes medical conditions observed, while the impression section is a one sentence conclusion or diagnosis.
Notice Spinks and Moens \cite{spinks2019justifying} filtered the \textit{findings} section, and kept only sentences referring to one disease (Cardiomegaly). The papers that seem to show the best performance in terms of NLP metrics are
KERP \cite{li2019knowledge}, CLARA \cite{biswal2020clara} and Xue et al. 2019 \cite{10.1007/978-3-030-20351-1_10} for the findings section,
MTMA \cite{10.1007/978-3-030-33850-3_8} for the impression section,
and Yuan et al. \cite{yuan2019automatic}, MLMA \cite{10.1007/978-981-15-4015-8_15} and Xue et al. 2019 \cite{10.1007/978-3-030-20351-1_10} for both sections.
\surveyUpdate{Of these, only MTMA has a large difference to its competitors, and there is no clear winner in the other sections}. 
\surveyUpdate{Some caveats, however, should be kept in mind when interpreting these results:}

    \surveyUpdate{(1) The results reported in the literature only allow comparisons in terms of standard natural language metrics (BLEU, ROUGE-L, etc.), but from these results we cannot draw conclusions about medical correctness, since NLP metrics and clinical accuracy are not necessarily correlated.}
    
    (2) MTMA uses additional input, as discussed in section \ref{section:input-output}.
    Specifically, the model receives the indication and findings sections of the report to generate the impression section, at both test and train stages.
    In a sense, this could be seen as an enhanced summarizing approach, since the impression section contains a conclusion from the findings.
    
    (3) Some NLP metrics, such as CIDEr, ROUGE-L and METEOR, have variants and parameters, as discussed in section \ref{section:metrics}.
    Unfortunately, most papers do not mention the specific version or implementation used.

    \surveyUpdate{(4) The IU X-ray dataset does not have standard training-validation-test splits. 
    This has led researchers to define their own splits, as indicated by column \textit{Split} of Table \ref{table:iu-xray-results}. These splits are not consistent across papers, making results less comparable. For example, if a model was evaluated in an easier test split, that would give it an unfair advantage over other models evaluated in harder test splits. Additionally, other decisions such the number of images per report (frontal, lateral or both), the tokenization algorithm employed, the removal of noisy sentences, the removal of words with a frequency under a given threshold, the removal of duplicate images, among other preprocessing decisions, are not always explicitly stated in papers, and these may have an impact on the results as well.}
    
    \surveyUpdate{(5) These are overall results only, so a more fine-grained performance assessment on specific abnormalities or diseases is missing. This further shows the need for standardizing one or more evaluation metrics to measure the medical correctness of a generated report, considering different aspects of interest.}


\newcommand{\tableSotaComment}[1]{$^{\textrm{(#1)}}$}

\def\addedColSep{2.5pt}

\addtolength{\tabcolsep}{-\addedColSep{}}

\begin{table}[tbp]
\scalebox{0.8}{
\begin{tabular}{|l|cccc|c|c|c|c|}
\hline
\textbf{Paper} & \textbf{BLEU-1} & \textbf{BLEU-2} & \textbf{BLEU-3} & \textbf{BLEU-4} & \textbf{ROUGE-L} & \textbf{METEOR} & \textbf{CIDEr-D} & \textbf{Split} \\ \hline
\multicolumn{9}{|c|}{\textbf{Findings section}} \\ \hline
Liu et al. \cite{liu2019clinically} & 0.369 & 0.246 & 0.171 & 0.115 & 0.359 & - & \textbf{1.490} & 7:1:2 $^{1}$ \\
HRGR \cite{li2018hybrid} & 0.438 & 0.298 & 0.208 & 0.151 & 0.369 & - & 0.343 & 7:1:2 $^{2}$ \\
KERP \cite{li2019knowledge} & \textbf{0.482} & \textbf{0.325} & \textbf{0.226} & 0.162 & 0.339 & - & 0.280 & 7:1:2 $^{2}$ \\
TieNet \cite{wang2018tienet} \tableSotaComment{1} & 0.330 & 0.194 & 0.124 & 0.081 & 0.311 & - & 1.334 & 7:1:2 $^{1}$ \\
Xue et al. 2018 \cite{xue2018multimodal} \tableSotaComment{2} & 0.441 & 0.320 & 0.231 & 0.181 & 0.366 & 0.220 & 0.343 & 2,525/250 \\
RTMIC \cite{xiong2019reinforced} & 0.350 & 0.234 & 0.143 & 0.096 & - & - & 0.323 & 7:2:1 \\
CLARA \cite{biswal2020clara} \tableSotaComment{3} & 0.471 & 0.324 & 0.214 & \textbf{0.199} & - & - & 0.359 & 7:1:2 \\
Xue et al. 2019 \cite{10.1007/978-3-030-20351-1_10} & 0.477 & 0.332 & 0.243 & 0.189 & \textbf{0.380} & \textbf{0.223} & 0.320 & 3,031/300 \\
CMAS \cite{jing2019show} & 0.464 & 0.301 & 0.210 & 0.154 & 0.362 & - & 0.275 & Unk \\ \hline
\multicolumn{9}{|c|}{\textbf{Findings section - Cardiomegaly sentences only}} \\ \hline
Spinks and Moens \cite{spinks2019justifying} & 0.490 & 0.350 & 0.250 & 0.180 & 0.400 & 0.270 & 0.600 & 8:1:1 \\
\hline
\multicolumn{9}{|c|}{\textbf{Impression section}} \\ \hline
MTMA \cite{10.1007/978-3-030-33850-3_8} & \textbf{0.882} & \textbf{0.874} & \textbf{0.867} & \textbf{0.860} & \textbf{0.929} & - & - & 5,461/500/500 \\
CMAS \cite{jing2019show} & 0.401 & 0.290 & 0.220 & 0.166 & 0.521 & - & 1.457 & Unk \\ \hline
\multicolumn{9}{|c|}{\textbf{Findings + Impression sections}} \\ \hline
CoAtt \cite{jing2017automatic} & 0.517 & \textbf{0.386} & 0.306 & 0.247 & 0.447 & 0.217 & 0.327 & 6,470/500/500 \\
Huang et al. \cite{huang2019multi} & 0.476 & 0.340 & 0.238 & 0.169 & 0.347 & - & 0.297 & 8:1:1 \\
Yuan et al. \cite{yuan2019automatic} & \textbf{0.529} & 0.372 & 0.315 & 0.255 & 0.453 & \textbf{0.343} & - & 8:2 \\
Xue et al. 2018 \cite{xue2018multimodal} & 0.464 & 0.358 & 0.270 & 0.195 & 0.366 & 0.274 & - & 2,775/250 \\
Vispi \cite{li2019vispi} & 0.419 & 0.280 & 0.201 & 0.150 & 0.371 & - & 0.553 & 7:1:2 \\
Singh et al. \cite{singh2019chest} & 0.374 & 0.224 & 0.153 & 0.110 & 0.308 & 0.164 & 0.360 & 6,718/350/350 \\
Yin et al. \cite{8970668} & 0.445 & 0.292 & 0.201 & 0.154 & 0.344 & 0.175 & 0.342 & 6,470/500/500 \\
MLMA \cite{10.1007/978-981-15-4015-8_15} & 0.500 & 0.380 & \textbf{0.317} & \textbf{0.278} & 0.440 & 0.281 & \textbf{1.067} & 6,429/500/500 \\
Harzig et al. 2019a \cite{harzig2019addressing} & 0.373 & 0.246 & 0.175 & 0.126 & 0.315 & 0.163 & 0.359 & 90:5:5 \\
A3FN \cite{10.1007/978-3-030-18590-9_64} & 0.443 & 0.337 & 0.236 & 0.181 & 0.347 & - & 0.374 & 9:1 \\
Xue et al. 2019 \cite{10.1007/978-3-030-20351-1_10} & 0.489 & 0.340 & 0.252 & 0.195 & \textbf{0.478} & 0.230 & 0.565 & 3,031/300 \\
Zhang et al. \cite{zhang2020radiology} & 0.441 & 0.291 & 0.203 & 0.147 & 0.367 & - & 0.304 & 5-fold CV \\
\hline
\end{tabular}
}
\caption{Evaluation results of papers that use the IU X-ray dataset.
    All values were extracted from their papers, except in some cases where results were not present in the own paper:
    \tableSotaComment{1} TieNet \cite{wang2018tienet} results were presented in Liu et al. \cite{liu2019clinically} as a baseline;
    \tableSotaComment{2} Xue et al. 2018 \cite{xue2018multimodal} results in the findings section were presented in Xue et al. 2019 \cite{10.1007/978-3-030-20351-1_10} as a baseline.
    \tableSotaComment{3} CLARA \cite{biswal2020clara} results are from the fully automatic version.
    }
\label{table:iu-xray-results}
\vspace{-8mm}
\end{table}

\addtolength{\tabcolsep}{\addedColSep{}}

\section{Challenges and future work}
\label{section:challenges}

In this section, we identify unsolved challenges in the literature and potential avenues for future research in the task of report generation from medical images.

\paragraph{Protocol for expert evaluation}
If the ultimate goal is to develop a report generation system that meets high-quality standards, 
it makes sense that such a system be thoroughly tested by medical experts to evaluate its performance in different clinical settings.
Most papers reviewed are weak in this regard, as only four
of them
\cite{gale2018producing, 10.1007/978-3-030-00934-2_78, 10.1007/978-3-030-32251-9_37, spinks2019justifying}
perform a correctness evaluation with medical experts,
meaning that 90\% of the works does not carry out an expert evaluation, feedback that should be immensely valuable to understand the strengths and weaknesses of a model.
Therefore, a clear avenue for improvement is to standardize a protocol for human evaluation of these systems by imaging experts, for example starting with chest X-rays, which is the medical image type with more datasets available and research done.
A standard protocol should facilitate fair comparisons between studies and allow to assess how close a model is to meet standard criteria for deployment in a clinical setting.

\surveyUpdate{The expertise of the human evaluators is an important factor to consider as well. It stands to reason that the judgement of a board-certified radiologist with years of experience should be more reliable than the judgement of a physician with limited experience. Similarly, the consensus of a team of radiologists should be preferred over a single radiologist. In the same line, measuring the inter-agreement of several radiologists can help to better assess the difficulty of the task itself \cite{jain2021radgraph}. If radiologists tend to disagree more, this may indicate an inherent ambiguity in the task that could be the explanation for the possible underperformance of a given model.}

\label{section:chall-metric-med-corr}
\paragraph{Automatic metrics for medical correctness}
Having a proper expert evaluation is desirable. However, it is not feasible to ask radiologists to manually evaluate hundreds of machine-generated reports every time a small tweak in a model is performed.
Instead, one would like to have one or more automatic metrics positively correlated with expert human evaluation, in order to speed up the model design and testing cycle.
We found that more than 70\% of the works reviewed (29 out of 40) limit the automatic report evaluation to traditional NLP metrics such as BLEU, ROUGE-L or CIDEr, which are not designed to evaluate a report from a medical correctness point of view \cite{boagbaselines,zhang2019optimizing, liu2019clinically,babar2021evaluating,pino2021clinically,pino2020inspecting}.
\surveyUpdate{Furthermore, these evaluation methods have been recently  contested in other NLP tasks \cite{van-miltenburg-etal-2020-gradations, van-miltenburg-etal-2021-underreporting, reiter2018structured, mathur2020tangled}.}
%
Some works tried to remedy this limitation by devising their own auxiliary metrics to evaluate medical correctness to some degree
\cite{liu2019clinically, huang2019multi, li2018hybrid, xue2018multimodal, 10.1007/978-3-030-32251-9_37, biswal2020clara, 10.1007/978-3-030-18590-9_64, zhang2020radiology, jing2019show, moradi2016cross, wu2017generative},
which are interesting approaches.
We highlight the metric MIRQI proposed very recently by Zhang et al. \cite{zhang2020radiology}, which is very similar to SPICE \cite{10.1007/978-3-319-46454-1_24} as described in section \ref{section:metrics:med-corr}, as it attempts to build a graph capturing abnormalities, their relations and attributes.
We believe this is the most sophisticated metric for medical correctness found in the literature, and great ideas can be adopted from it.

Unfortunately, all the proposed metrics lack validation by medical experts, as none of the papers presents the results of a study assessing the correlation between the proposed metric and expert medical judgment. Thus, finding one or more golden automatic metrics for medical correctness remains an open problem.
To solve it, the precision and accuracy of a report is critical in the medical domain and need to be captured \cite{zhang2019optimizing, babar2021evaluating,pino2021clinically}, whereas other aspects such as natural language fluency should probably weigh less in importance.
We believe designing and validating such metrics is a clear avenue for future research, with the potential to have a significant impact on the field.

\paragraph{Improve explainability}
To build trust in an AI system, a desirable feature is the ability to provide clear and coherent explanations for its decisions \cite{8466590, 8400040}.
This is particularly relevant in the healthcare domain, where decisions have to be made with extreme caution since the patient's health is at stake.
Thus, high levels of transparency, interpretability, and accountability are required to justify the outputs delivered, align to the expert's expectations, and acquire their trust \cite{tonekaboni2019clinicians, tjoa2019survey, reyes2020interpretability}.


Only two papers reviewed \cite{gale2018producing, spinks2019justifying} have explainability as a primary focus, as discussed in section \ref{section:xai}, though one of them \cite{gale2018producing} does not provide an explanation for the report.
Additionally, some works mention some form of local explainability in their models, but always as a secondary output and giving it a rather superficial treatment, with no rigorous evaluation.
In the absence of empirical results across all papers, we cannot draw conclusions about which explanation techniques are better or worse.
Thus, a potential avenue for future research is explainability with a more rigorous and empirical focus, and possibly including other approaches, such as global explanations, uncertainty, or more, which may be necessary for clinicians \cite{tonekaboni2019clinicians}.
We believe this research avenue will benefit from the feedback and evaluation of medical imaging experts, who are the end-users of these systems.
What would be a suitable explanation for a radiologist?
In a multi-sentence report, how should the explanation be structured?
An expert's opinion is valuable for answering these and other questions, and ultimately for assessing the explanation.

\paragraph{New learning strategies and architectures.}
If the ultimate goal is to have a model that learns to generate accurate and useful medical reports, the optimization strategy employed should be designed to guide the model in this direction.
As we saw in section \ref{section:opt-strat}, most papers used teacher-forcing, a training strategy which is domain-agnostic and thus suboptimal for the medical domain \cite{zhang2019optimizing}. Similarly, a few papers used reinforcement learning \cite{liu2019clinically, li2018hybrid, xiong2019reinforced, 10.1145/3357254.3357256, jing2019show} with traditional NLP metrics as rewards, which are not designed for medicine either. Only Liu et al. \cite{liu2019clinically} included a domain-specific reward that explicitly promotes medical correctness. Unfortunately, a manual inspection of several generated reports conducted by the authors revealed that the model was missing positive findings (low recall) as well as failing to provide accurate descriptions of the positive findings detected.

Given these reasons, there is still room for finding better optimization strategies for image-based medical report generation. In this regard, reinforcement learning appears to be the most promising training paradigm to explore, as illustrated by the work of Zhang et al. \cite{zhang2019optimizing} on factual correctness optimization in a related medical task. If a robust medical correctness metric is developed (as previously discussed in this section), then the metric could be used as a reward in a reinforcement learning setting to teach the model to generate reports that are medically correct.

\paragraph{Other image modalities and body regions are less explored}
Most research has concentrated on chest X-rays, as 24 out of 40 papers focus their study on this image type.
This modality presents a very specific nature and different characteristics from other imaging studies.
For example, when a radiologist reads a chest X-ray, the focus is on the underlying anatomy and identification of possible areas of distortion based on different densities of the image.
On the other hand, when analyzing a PET image, the focus is on detecting areas of increased radiotracer activity; for MRI scans, the radiologist may review several images obtained with different configurations at the same time; and for each other modality there may be more specific conditions.
Hence, the results shown here are highly biased towards chest X-rays, which will not necessarily extrapolate to other scenarios.

Notice there are datasets with multiple image types or body parts, namely ImageCLEF caption \cite{eickhoff2017overview, de2018overview}, ROCO \cite{dataset2018roco} and PEIR Gross \cite{jing2017automatic}, as it was mentioned in section \ref{section:datasets}.
However, we believe their broad nature, i.e., the inclusion of many types and regions simultaneously, may be a drawback when trying to apply an advanced deep learning approach, for three main reasons.
First, it is more difficult to include specific domain knowledge in the models, as the knowledge should cover all modalities and body parts.
Second, assessing medical correctness is more complicated, since domain knowledge is needed to design these metrics, as noted in section \ref{section:metrics}.
Third, it would be more challenging to provide interpretability for the model, as the explanations should cover all modalities.
Ultimately, we believe better solutions can be achieved by designing them for a specific problem and setting.
In conclusion, there is a clear opportunity to extend research into other image types and body regions by raising new collections with other image types, evaluating the same methods in different modalities, or further covering the existing datasets.

\paragraph{Explore more alternatives to include domain knowledge}

As we saw in section \ref{section:domain-knowledge}, the approaches explored in the literature for incorporating domain knowledge into models are (1) the use of graph neural networks at the visual component level and (2) the use template databases curated with expert knowledge---in addition to the widespread use of auxiliary tasks, that can be viewed as a way of domain knowledge transfer as well.
However, other approaches remain unexplored. A recent survey by Xie et al. \cite{xie2020survey} synthesizing over 270 papers on domain knowledge for deep learning-based medical image analysis presents interesting ideas that could be applicable to the report generation setting. For example, curriculum learning \cite{bengio2009curriculum} and self-paced learning \cite{kumar2010self} could be used to imitate the learning curve from easier to harder instances that radiologists go through when they learn to interpret and diagnose images. Also, the use of handcrafted algorithms to extract visual features that better capture what radiologists focus on in an image could be used, which many works have verified to have synergistic effects in combination with the features learned by the CNN \cite{xie2020survey}. This would improve the quality of the visual component and potentially translate into better reports. Studying how imaging experts analyze an image, how they focus the attention to different regions of the image as needed, could be useful to inspire innovations in model architectures in order to emulate that process.

\paragraph{Medical Human-AI interaction}
Most reviewed works leave aside important aspects pertaining the model's integration in a real clinical setting and its interaction with clinicians as an AI assistant.
Besides high levels of accuracy, there are other needs a system should aim to meet in a medical human-AI collaboration workflow.
For example, Cai et al. \cite{cai2019hello} argue that clinicians should have transparent information about the model's overall strengths and weaknesses, its subjective point-of-view, its overall design objective and how exactly it uses the information to derive a final diagnosis.
Also, Amershi et al. \cite{amershi2019guidelines} proposed and validated several design guidelines for general human-AI interaction that can be relevant in the context of automatic report generation, such as \textit{Make clear why the system did what it did} via explanations, and \textit{Support efficient correction} by making it easy to edit, refine, or recover when the AI system is wrong.
Among all papers reviewed, only CLARA \cite{biswal2020clara} targets an explicit workflow with human interaction, in which a report is generated cooperatively by a human who types some preliminary text and the system autocompletes the rest.

Also, there are potential use cases that an AI assistant for report generation can face in routine practice which are not addressed in the reviewed literature.
For example, (1) \textit{open-ended visual question answering (VQA)}: instead of a full report with too many details, a clinician might be interested in the model's opinion on a specific aspect of the image(s).
This query could be expressed as a natural language question that the model would have to answer, which would require a model with open-ended VQA capabilities.
Although this is a different task than report generation, we believe the latter could be approached as giving answers to a sequence of questions from physicians, allowing a richer interaction between the expert and the system.
The multiple ImageCLEF challenges involving a medical VQA task \cite{ImageCLEFVQA-Med2018, ImageCLEFVQA-Med2019, ImageCLEF-VQA-Med2020} and the recently published PathVQA dataset \cite{he2020pathvqa} could be helpful in exploring this direction.
(2) \textit{Reporting temporal information}: sometimes clinicians are interested in the evolution of a health condition by analyzing a sequence of imaging snapshots over time, rather than describing a single image.
None of the surveyed papers considers this use case.
(3) \textit{Quantitative Radiology}: in some cases a clinician might be interested in specific numerical measurements to further assess the patient's condition, for example, the degree of a certain property in the tissues \cite{jackson2018quantitative}.
This adds more complexity to the problem, since models would need the ability to make these accurate numerical measurements, in addition to interpreting them through words in the generated report.
In sum, there may be different ways to fulfill the report generation task, and we believe researchers should aim to find the most useful approaches for clinicians in each specific environment.

\section{Limitations}
\label{section:limitations}

The main limitations of this survey are two.
First, new papers on report generation from medical images are published relatively often, we tried to be as comprehensive as possible and include all of them, but we do not rule out that some papers may have been missed.
Second, we left out of the analysis works from related tasks, such as disease classification, report summarizing, or medical image segmentation.
These topics may have interesting approaches or insights on how to improve the visual features generated, how to optimize the text generation, evaluation techniques, and more.

\section{Conclusions}
\label{section:conclusion}

In this work, we have reviewed the state of research in deep-learning based methods for automatic report generation from medical images, in terms of different key aspects.
First, we described the report and classification \textbf{datasets} available and commonly used in the literature, totalling 27 collections, which cover different image modalities and body parts, and include useful tags and localization information.
Second, we presented an analysis of \textbf{model designs} in terms of standard practices, inputs and outputs, visual components, language components, domain knowledge, auxiliary tasks, and optimization strategies.
We cannot recommend an optimal model design due to the lack of proper evaluations, but several guidelines can be inferred.
For instance, a robust visual component should make use of CNNs and would certainly benefit from training in auxiliary medical image tasks. Also, complementing the visual input with semantic information via tags or input text (e.g the report's \textit{indication} section) or access to a template database generally improves the language component's performance. Multitask learning to integrate the supervision of multiple tasks and reinforcement learning to directly optimize for factual correctness or other metrics of interest in generated reports appear as the most promising optimization approaches.
Third, we analyzed the \textbf{interpretablity} approaches employed in the literature, and found that many models provide a secondary output that can be used as a local explanation, either by providing a feature importance map, a counter-factual example, or by increasing the system's transparency.
However, only two works focused explicitly on studying this concern, by discussing extensively and providing formal evaluations.
Additionally, many other approaches can be explored, and hence this remains a heavily understudied aspect of this task.
Fourth, we discussed usual practices regarding \textbf{evaluation metrics}, and we found that most models are only evaluated
with traditional n-gram based NLP metrics not designed for medicine,
which are not able to capture the essential medical facts in a written report.
Next, we presented a comparison of papers' \textbf{performance results} on IU X-Ray, the most frequently used dataset, but limited to said NLP metrics that papers report,
making us unable to judge models from a medical perspective.

Lastly, we identified \textbf{challenges} in the field that none of the reviewed papers has successfully addressed, and we proposed avenues for future research where we believe possible solutions could be found.
The main challenges lay on improving the evaluation methods employed, by developing a \textit{standard protocol for expert evaluation} and \textit{automatic metrics for medical correctness}.
Other important aspects are improving the \textit{explainability of models}, and considering the \textit{medical human-AI interaction}.
We intend this survey to serve as an entry point for researchers who want an overview of the current advances in the field and also to raise awareness of critical problems that future research should focus on, with the end goal of developing mature and robust technologies that can bring value to healthcare professionals and patients in real clinical settings.

\bibliographystyle{ACM-Reference-Format}
\bibliography{sample-base}

\newpage
\section{Supplementary Material}

\subsection{Datasets}
\label{section:appendix:datasets}

Next, we include Table \ref{table:datasets-amounts-full} with the main highlights of all datasets, including both public and proprietary, and Table \ref{table:datasets-extra-data-full} with details of the additional information provided by each collection.

\newcommand{\datasetSpanishAppendix}{$^{\textrm{(sp)}}$}
\newcommand{\datasetPortugueseAppendix}{$^{\textrm{(pt)}}$}
\newcommand{\datasetChineseAppendix}{$^{\textrm{(ch)}}$}
\newcommand{\datasetPendingReleaseAppendix}{$^{\textrm{(1)}}$}
\newcommand{\datasetPMCAppendix}{$^{\textrm{(2)}}$}
\newcommand{\datasetROCOAppendix}{$^{\textrm{(3)}}$}
\newcommand{\datasetVideoFramesAppendix}{$^{\textrm{(4)}}$}
\newcommand{\datasetNoneAppendix}{$^{\textrm{(5)}}$}

\begin{table}[!htb]
\scalebox{0.78}{
\begin{tabular}{|p{0.25\linewidth}|c|p{0.27\linewidth}|c|c|c|p{0.18\linewidth}|}
\hline
\textbf{Dataset} & \textbf{Year} & \textbf{Image Type} & \textbf{\# images} & \textbf{\# reports} & \textbf{\# patients} & \textbf{Used by papers} \\ \hline
\multicolumn{7}{|c|}{\textbf{Public report datasets}} \\ \hline
IU X-ray \cite{10.1093/jamia/ocv080} & 2015 & Chest X-Ray & 7,470 & 3,955 & 3,955 & \cite{jing2017automatic, liu2019clinically, huang2019multi, yuan2019automatic, li2018hybrid, li2019knowledge, wang2018tienet, xue2018multimodal, li2019vispi, xiong2019reinforced, singh2019chest, 8970668, 10.1007/978-3-030-33850-3_8, gasimova2019automated, 10.1007/978-981-15-4015-8_15, harzig2019addressing, biswal2020clara, 10.1007/978-3-030-18590-9_64, 10.1007/978-3-030-20351-1_10, zhang2020radiology, jing2019show, Shin_2016_CVPR} \\ \hline
MIMIC-CXR \cite{johnson2019mimicv1, johnson2019mimiccxrjpg} & 2019 & Chest X-Ray & 377,110 & 227,827 & 65,379 & \cite{liu2019clinically} \\ \hline
PadChest\datasetSpanish \cite{bustos2019padchest} & 2019 & Chest X-Ray & 160,868 & 109,931 & 67,625 & None\datasetNone \\ \hline
\raggedright{ImageCLEF Caption 2017 \cite{eickhoff2017overview} } & 2017 & Biomedical\datasetPMC & 184,614 & 184,614 & - & \cite{10.1007/978-3-319-98932-7_21} \\ \hline
\raggedright{ImageCLEF Caption 2018 \cite{de2018overview} } & 2018 & Biomedical\datasetPMC & 232,305 & 232,305 & - & None\datasetNone \\ \hline 
ROCO \cite{dataset2018roco} & 2018 & Multiple radiology\datasetROCO & 81,825 & 81,825 & - & None\datasetNone \\ \hline
PEIR Gross \cite{jing2017automatic} & 2017 & Gross lesions & 7,442 & 7,442 & - & \cite{jing2017automatic} \\ \hline
INBreast\datasetPortuguese \cite{moreira2012inbreast} & 2012 & Mammography X-ray & 410 & 115 & 115 & \cite{10.1007/978-3-030-26763-6_66, 10.1145/3357254.3357256} \\ \hline
STARE \cite{hoover1975stare} & 1975 & Retinal fundus & 400 & 400 & - & None\datasetNone \\ \hline 
RDIF\datasetPendingRelease \cite{maksoud2019coral8} & 2019 & Kidney Biopsy & 1,152 & 144 & 144 & \cite{maksoud2019coral8} \\ \hline \hline
\multicolumn{7}{|c|}{\textbf{Private report datasets}} \\ \hline
CX-CHR\datasetChinese \cite{li2018hybrid, li2019knowledge, jing2019show} & 2018 & Chest X-Ray & 45,598 & 35,609 & 35,609 & \cite{li2018hybrid, li2019knowledge, jing2019show} \\ \hline
TJU\datasetChinese \cite{8935910} & 2019 & Chest X-Ray & 19,985 & 19,985 & - & \cite{8935910} \\ \hline
Hip fracture \cite{gale2017detecting, gale2018producing} & 2017 & Hip X-Ray & 53,279 & 4,010 & 26,639 & \cite{gale2018producing} \\ \hline
Ultrasound \cite{Zeng2018, zeng2020deep} & 2018 & \raggedright{Gallbladder, kidney and liver ultrasound\datasetVideoFrames } & 4,302 & 4,302 & - & \cite{Zeng2018, zeng2020deep} \\ \hline
Fetal Ultrasound \cite{10.1007/978-3-030-32251-9_37} & 2019 & Fetal ultrasound\datasetVideoFrames & 2,800 & 2,800 & - & \cite{10.1007/978-3-030-32251-9_37} \\ \hline
CINDRAL \cite{10.1007/978-3-030-04224-0_24} & 2018 & Cervical neoplasm WSI & 1,000 & 1,000 & 50 & \cite{10.1007/978-3-030-04224-0_24} \\ \hline
BCIDR \cite{zhang2017mdnet} & 2017 & Bladder biopsy & 1,000 & 1,000 & 32 & \cite{zhang2017mdnet} \\ \hline
Continuous wave \cite{moradi2016cross} & 2016 & \raggedright{Continuous wave doppler echocardiography} & 722 & 10,479 & - & \cite{moradi2016cross} \\ \hline
\hline
\multicolumn{7}{|c|}{\textbf{Public classification datasets}} \\ \hline
CheXpert \cite{irvin2019chexpert} & 2019 & Chest X-Ray & 224,316 & 0 & 65,240 & \cite{yuan2019automatic, zhang2020radiology} \\ \hline
ChestX-ray14 \cite{Wang_2017_CVPR} & 2017 & Chest X-Ray & 112,120 & 0 & 30,805 & \cite{li2019knowledge, wang2018tienet, li2019vispi, xiong2019reinforced, biswal2020clara, 10.1007/978-3-030-20351-1_10, jing2019show} \\ \hline
LiTS \cite{christ2017lits} & 2017 & Liver CT scans & 200 & 0 & - & \cite{10.1007/978-3-030-00934-2_78} \\ \hline 
\raggedright{ACM Biomedia 2019 \cite{10.1145/3343031.3356058}} & 2019 & Gastrointestinal tract \datasetVideoFrames & 14,033 & 0 & - & \cite{10.1145/3343031.3356066} \\ \hline
DIARETDB0 \cite{kauppi2006diaretdb0} & 2006 & Retinal fundus & 130 & 0 & - & \cite{wu2017generative} \\ \hline
DIARETDB1 \cite{kalviainen2007diaretdb1} & 2007 & Retinal fundus & 89 & 0 & - & \cite{wu2017generative} \\ \hline
Messidor \cite{decenciere2014feedback, abramoff2013automated} & 2013 & Retinal fundus & 1,748 & 0 & 874 & \cite{wu2017generative} \\ \hline
DDSM \cite{heath2001digital} & 2001 & Mammography X-ray & 10,480 & 0 & - & \cite{kisilev2016medical} \\ \hline
\hline
\multicolumn{7}{|c|}{\textbf{Private classification datasets}} \\ \hline
MRI Spine \cite{10.1007/978-3-030-00937-3_22} & 2018 & Spine MRI scans & $\geq$253 & 0 & 253 & \cite{10.1007/978-3-030-00937-3_22} \\ \hline
\end{tabular}
}
\caption{Datasets used in the literature.
    WSI stands for Whole Slide Images.
    All reports are written in English, except those marked with \datasetSpanishAppendix{} which are in Spanish, with \datasetChineseAppendix{} in Chinese, and \datasetPortugueseAppendix{} in Portuguese.
    Other notes, 
    \datasetPendingReleaseAppendix: the RDIF dataset is pending release.
    \datasetPMCAppendix: for the ImageCLEF datasets, images were extracted from PubMed Central papers and filtered with an automatically to keep only clinical images, then it contains samples from other domains.
    \datasetROCOAppendix: contains multiple modalities, namely CT, Ultrasound, X-Ray, Fluoroscopy, PET, Mammography, MRI, Angiography and PET-CT.
    \datasetVideoFramesAppendix: the images are frames extracted from videos.
    \datasetNoneAppendix: none of the papers reviewed used this dataset.
    }
\label{table:datasets-amounts-full}
\vspace{-6mm}
\end{table}

\newcommand{\datasetCellEmptyAppendix}{\multicolumn{1}{c|}{-}}

\def\addedColSepDatasets{1pt}

\addtolength{\tabcolsep}{-\addedColSepDatasets{}}

\begin{table}[!htb]
\scalebox{0.8}{
\begin{tabular}{|p{0.25\linewidth}|p{0.1\linewidth}|p{0.32\linewidth}|p{0.22\linewidth}|p{0.25\linewidth}|}
\hline
\textbf{Dataset} & \textbf{Text} & \textbf{Tags} & \raggedright{\textbf{Tags annotation method}} & \textbf{Localization} \\ \hline
\multicolumn{5}{|c|}{\textbf{Public report datasets}} \\ \hline
IU X-ray \cite{10.1093/jamia/ocv080} & Indication & (1) MeSH and RadLex concepts (2) MeSH concepts & \raggedright{(1) Manual (2) MTI and MetaMap} & \datasetCellEmptyAppendix \tabularnewline \hline
MIMIC-CXR \cite{johnson2019mimicv1, johnson2019mimiccxrjpg} & Comparis, Indicat & 14 CheXpert labels & CheXpert labeler and NegBio & \datasetCellEmptyAppendix \tabularnewline \hline
PadChest \cite{bustos2019padchest} & \datasetCellEmptyAppendix & 297 labels (findings, diagnoses and anatomic) & \raggedright{27\% manual, rest by RNN} & \datasetCellEmptyAppendix \tabularnewline \hline
\raggedright{ImageCLEF Caption 2017 \cite{eickhoff2017overview}} & \datasetCellEmptyAppendix & UMLS tags & Quick-UMLS & \datasetCellEmptyAppendix \\ \hline
\raggedright{ImageCLEF Caption 2018 \cite{de2018overview}} & \datasetCellEmptyAppendix & UMLS tags & Quick-UMLS & \datasetCellEmptyAppendix \\ \hline
ROCO \cite{dataset2018roco} & \datasetCellEmptyAppendix & UMLS tags & Quick-UMLS & \datasetCellEmptyAppendix \\ \hline
PEIR Gross \cite{jing2017automatic} & \datasetCellEmptyAppendix & Top words & Top TF-IDF scores & \datasetCellEmptyAppendix \\ \hline
INBreast \cite{moreira2012inbreast} & \datasetCellEmptyAppendix & Abnormalities & Manual & \raggedright{Abnormality contours} \tabularnewline \hline
STARE \cite{hoover1975stare} & \datasetCellEmptyAppendix & \raggedright{Levels for 39 conditions and presence of 13 diagnostics} & Manual & \datasetCellEmptyAppendix \\ \hline
RDIF \cite{maksoud2019coral8} & Indication & \datasetCellEmptyAppendix & \datasetCellEmptyAppendix & \datasetCellEmptyAppendix \\ \hline\hline
\multicolumn{5}{|c|}{\textbf{Private report datasets}} \\ \hline
CX-CHR \cite{li2018hybrid, li2019knowledge, jing2019show} & \datasetCellEmptyAppendix & \datasetCellEmptyAppendix & \datasetCellEmptyAppendix & \datasetCellEmptyAppendix \\ \hline
TJU \cite{8935910} & \datasetCellEmptyAppendix & Top abnormality words & \raggedright{40 most frequent words} & \datasetCellEmptyAppendix \\ \hline
Hip fracture \cite{gale2017detecting, gale2018producing} & \datasetCellEmptyAppendix & \raggedright{(1) Fracture presence, (2) fracture location and character} & \raggedright{(1) CNN \cite{gale2017detecting}, (2) manual \cite{gale2018producing}} & \datasetCellEmptyAppendix \\ \hline
Ultrasound \cite{Zeng2018, zeng2020deep} & \datasetCellEmptyAppendix & \raggedright{Organ and disease} & Manual & Organ bounding boxes \\ \hline
Fetal Ultrasound \cite{10.1007/978-3-030-32251-9_37} & \datasetCellEmptyAppendix & Body part & Manual & \datasetCellEmptyAppendix \\ \hline
CINDRAL \cite{10.1007/978-3-030-04224-0_24} & \datasetCellEmptyAppendix & Severity level for 4 attributes and diagnosis label & Manual & \datasetCellEmptyAppendix \\ \hline
BCIDR \cite{zhang2017mdnet} & \datasetCellEmptyAppendix & Disease status (4 possible) & Manual & \datasetCellEmptyAppendix \\ \hline
\raggedright{Continuous wave \cite{moradi2016cross}} & \datasetCellEmptyAppendix & Valve types & Manual & \datasetCellEmptyAppendix \\ \hline
\hline
\multicolumn{5}{|c|}{\textbf{Public classification datasets}} \\ \hline
CheXpert \cite{irvin2019chexpert} & \datasetCellEmptyAppendix & 14 CheXpert labels & CheXpert labeler & \datasetCellEmptyAppendix \\ \hline
ChestX-ray14 \cite{Wang_2017_CVPR} & \datasetCellEmptyAppendix & 14 disease labels & DNorm and MetaMap & \raggedright{Disease bounding boxes for 880 images} \tabularnewline \hline
LiTS \cite{christ2017lits} & \datasetCellEmptyAppendix & \datasetCellEmptyAppendix & \datasetCellEmptyAppendix & \raggedright{Liver and tumor segmentation masks} \tabularnewline \hline
\raggedright{Gastrointestinal challenge \cite{10.1145/3343031.3356058}} & \datasetCellEmptyAppendix & 16 labels (e.g. anatomic, pathological or surgery findings) & Manual & \datasetCellEmptyAppendix \\ \hline
DIARETDB0 \cite{kauppi2006diaretdb0} & \datasetCellEmptyAppendix & DR severity level & Manual & Abnormality contours \\ \hline
DIARETDB1 \cite{kalviainen2007diaretdb1} & \datasetCellEmptyAppendix & DR severity level & Manual & Abnormality contours \\ \hline
Messidor \cite{decenciere2014feedback, abramoff2013automated} & \datasetCellEmptyAppendix & DR severity level & Manual & \datasetCellEmptyAppendix \\ \hline
DDSM \cite{heath2001digital} & \datasetCellEmptyAppendix & Density level & Manual & \raggedright{Abnormalities at pixel level} \tabularnewline \hline
\hline
\multicolumn{5}{|c|}{\textbf{Private classification datasets}} \\ \hline
MRI Spine \cite{10.1007/978-3-030-00937-3_22} & \datasetCellEmptyAppendix & \datasetCellEmptyAppendix & \datasetCellEmptyAppendix & \raggedright{Diseases and body parts at pixel level} \tabularnewline \hline
\end{tabular}
}
\caption{Additional data contained in each dataset.
    MeSH \cite{rogers1963medical} and RadLex \cite{langlotz2006radlex} are sets of medical concepts.
    MTI \cite{mork2013nlm}, MetaMap \cite{aronson2010overview}, CheXpert labeler \cite{irvin2019chexpert}, NegBio \cite{peng2018negbio}, Quick-UMLS \cite{soldaini2016quickumls} and DNorm \cite{leaman2015challenges} are automatic labeler tools.
    \textit{Manual} means manually annotated by experts.
    In all cases, the localization information was manually annotated by experts.
    }
\label{table:datasets-extra-data-full}
\end{table}

\addtolength{\tabcolsep}{\addedColSepDatasets{}}

\FloatBarrier

\clearpage
\subsection{Auxiliary Tasks}
\label{section:appendix:auxtasks}
Table \ref{tab:aux-tasks} presents the categories of auxiliary tasks identified in the literature and which papers implemented them.

\begin{table}[ht]
\scalebox{0.8}{
\begin{tabular}{|p{0.5\textwidth}|p{0.5\textwidth}|}
\hline
 \textbf{Auxiliary Task}  & \textbf{Used by papers}
 \\ \hline
 Multi-label classification &
 \cite{jing2017automatic, yuan2019automatic, li2019knowledge, wang2018tienet, li2019vispi, xiong2019reinforced, 8935910, 8970668, 10.1007/978-3-030-33850-3_8, harzig2019addressing, biswal2020clara, 10.1145/3343031.3356066, 10.1007/978-3-030-26763-6_66, zhang2020radiology, jing2019show, Shin_2016_CVPR}\\
 Single-label classification &
 \cite{zhang2017mdnet, gale2018producing, 10.1007/978-3-030-04224-0_24, 10.1007/978-3-030-32251-9_37, gasimova2019automated, Zeng2018, 10.1007/978-3-319-98932-7_21, kisilev2016medical, moradi2016cross, spinks2019justifying, zeng2020deep}\\
 Sentence classification (normal/abnormal/stop) &
 \cite{harzig2019addressing, 10.1007/978-3-030-18590-9_64, jing2019show}\\
 Segmentation &
 \cite{10.1007/978-3-030-00934-2_78, 10.1007/978-3-030-00937-3_22}\\
 Object detection &
 \cite{kisilev2016medical, zeng2020deep}\\
 Attention weights regularization &
 \cite{maksoud2019coral8, 8970668}\\
 Embedding-to-embedding matching &
 \cite{8970668, moradi2016cross}\\
 Doc2vec & \cite{moradi2016cross}\\
 Text autoencoder &
 \cite{10.1007/978-3-030-33850-3_8, spinks2019justifying}\\
 GAN cycle-consistency & \cite{spinks2019justifying}\\
 \hline
\end{tabular}
}
\caption{Summary of auxiliary tasks used in the literature.}
\label{tab:aux-tasks}
\end{table}

\subsection{Optimization Strategies}
\label{section:appendix:optimization}
Table \ref{table:opti} presents the categories of optimization strategies identified in the literature and which papers implemented them.

\begin{table}[htbp]
\scalebox{0.8}{
\begin{tabular}{|p{0.25\textwidth}|p{0.4\textwidth}|p{0.5\textwidth}|}
\hline
 \textbf{Category} & \textbf{Optimization Strategy} & \textbf{Used by papers}
 \\ \hline
 
 \multirow[t]{3}{\linewidth}{Visual Component}
 & Pretrain in ImageNet & \cite{huang2019multi, li2018hybrid, wang2018tienet, xue2018multimodal, li2019vispi, singh2019chest, maksoud2019coral8, 8970668, 10.1007/978-3-030-04224-0_24, 10.1007/978-3-030-32251-9_37, gasimova2019automated, Zeng2018, 10.1007/978-3-030-20351-1_10, jing2019show, 10.1007/978-3-319-98932-7_21, moradi2016cross, zeng2020deep}\\ 
 & Train in Auxiliary Medical Image Tasks & \cite{yuan2019automatic, li2019knowledge, zhang2017mdnet, li2019vispi, xiong2019reinforced, gale2018producing, 10.1007/978-3-030-00934-2_78, 8935910, 8970668, 10.1007/978-3-030-33850-3_8, 10.1007/978-3-030-04224-0_24, 10.1007/978-3-030-32251-9_37, harzig2019addressing, biswal2020clara, Zeng2018, 10.1145/3343031.3356066, 10.1007/978-3-030-26763-6_66, zhang2020radiology, 10.1007/978-3-030-00937-3_22, jing2019show, Shin_2016_CVPR, 10.1007/978-3-319-98932-7_21, kisilev2016medical, zeng2020deep} \\
 & Train in Report Generation (end-to-end) & \cite{jing2017automatic, liu2019clinically, wang2018tienet, zhang2017mdnet, 10.1007/978-3-030-00934-2_78, 8970668, gasimova2019automated, harzig2019addressing, 10.1007/978-3-030-20351-1_10, 10.1145/3357254.3357256, 10.1007/978-3-319-98932-7_21, 10.1007/978-3-030-33850-3_8} \\
 \hline
 
 \multirow[t]{2}{\linewidth}{Report Generation}
 & Teacher-forcing & \cite{jing2017automatic, huang2019multi, yuan2019automatic, li2019knowledge, wang2018tienet, xue2018multimodal, zhang2017mdnet, li2019vispi, singh2019chest, maksoud2019coral8, gale2018producing, 10.1007/978-3-030-00934-2_78, 8935910, 8970668, 10.1007/978-3-030-33850-3_8, 10.1007/978-3-030-32251-9_37, gasimova2019automated, 10.1007/978-981-15-4015-8_15, harzig2019addressing, biswal2020clara, 10.1007/978-3-030-18590-9_64, Zeng2018, 10.1007/978-3-030-20351-1_10, 10.1007/978-3-030-26763-6_66, zhang2020radiology, 10.1145/3357254.3357256, jing2019show, Shin_2016_CVPR, 10.1007/978-3-319-98932-7_21, wu2017generative, spinks2019justifying, zeng2020deep}\\ 
 & Reinforcement Learning & \cite{liu2019clinically, li2018hybrid, xiong2019reinforced, 10.1145/3357254.3357256, jing2019show} \\
 \hline
 
 \multirow[t]{5}{\linewidth}{Other Losses or Training Strategies}
 & Multitask learning & \cite{jing2017automatic, li2019knowledge, wang2018tienet, zhang2017mdnet, maksoud2019coral8, 10.1007/978-3-030-00934-2_78, 8970668, 10.1007/978-3-030-33850-3_8, 10.1007/978-3-030-04224-0_24, harzig2019addressing, jing2019show, kisilev2016medical, spinks2019justifying, zeng2020deep}\\
 & Attention weights regularization & \cite{maksoud2019coral8, 8970668}\\
 & Contrastive loss & \cite{8970668} \\
 & Regression loss & \cite{kisilev2016medical, moradi2016cross, zeng2020deep} \\
 & Autoencoder & \cite{10.1007/978-3-030-33850-3_8, spinks2019justifying} \\
 & GAN & \cite{10.1007/978-3-030-00937-3_22, 10.1145/3357254.3357256, spinks2019justifying} \\
 \hline
\end{tabular}
}
\caption{Summary of optimization strategies used in the literature.}
\label{table:opti}
\end{table}







\end{document}